\documentclass[journal]{IEEEtran}

\usepackage{cite}

\usepackage{graphicx}
\usepackage{multirow}
\usepackage[table,xcdraw]{xcolor}
\usepackage{xcolor} 
\usepackage{float}
\usepackage{soul}   

\definecolor{fig9BlueText}{HTML}{8080FF} 
\definecolor{fig9PinkText}{HTML}{FF00FF} 
\definecolor{fig9GreenText}{HTML}{7C9A68} 

\definecolor{fig10OrangeHighlight}{HTML}{F7B68F} 
\definecolor{fig10GreenHighlight}{HTML}{77DD77} 
\definecolor{fig10BlueHighlight}{HTML}{9BC4E2} 

\newcommand{\colorBlue}[1]{{\color{fig9BlueText}#1}}
\newcommand{\colorPink}[1]{{\color{fig9PinkText}#1}}
\newcommand{\colorGreen}[1]{{\color{fig9GreenText}#1}}

\newcommand{\hlfigorange}[1]{\sethlcolor{fig10OrangeHighlight}\hl{#1}}
\newcommand{\hlfiggreen}[1]{\sethlcolor{fig10GreenHighlight}\hl{#1}}
\newcommand{\hlfigblue}[1]{\sethlcolor{fig10BlueHighlight}\hl{#1}}

\usepackage{amsmath}
\usepackage{amsthm}
\usepackage{amssymb}
\usepackage{booktabs}

\usepackage{siunitx}
\usepackage{array}
\usepackage{enumitem}
\usepackage{tabularx}
\usepackage{url}

\usepackage[caption=false,font=normalsize,labelfont=sf,textfont=sf]{subfig}
\ifCLASSINFOpdf
\else
\fi

\begin{document}
\title{Driving Style Recognition \textit{Like} an Expert Using Semantic Privileged Information from Large Language Models}

\author{    
        Zhaokun~Chen, 
        Chaopeng~Zhang,~\IEEEmembership{Student Member,~IEEE,}      
        Xiaohan~Li,
        Wenshuo~Wang,~\IEEEmembership{Member,~IEEE,}        Gentiane~Venture,~\IEEEmembership{Senior Member, IEEE}
        ~and~Junqiang~Xi
\thanks{This work was supported by the National Natural Science Foundation of China under Grant 52572469 and 52272411.  \textit{(Corresponding Authors: Wenshuo Wang and Junqiang Xi).}}
\thanks{Zhaokun Chen, Chaopeng Zhang and Xiaohan Li are with the School of Mechanical Engineering, Beijing Institute of Technology, Beijing, China (e-mail: zk.chen@bit.edu.cn;   cpzhang@bit.edu.cn; lixh030114@gmail.com).}

\thanks{Wenshuo Wang, and Junqiang Xi are with the Division of Energy-Mobility Convergence, Beijing Institute of Technology, Zhuhai, China (e-mail: ws.wang@bit.edu.cn; xijunqiang@bit.edu.cn).}
        
\thanks{Gentiane Venture is with the Department of Mechanical Engineering, The University of Tokyo, Tokyo, Japan (e-mail: venture@g.ecc.u-tokyo.ac.jp).}

}

\markboth{ACCEPTED FOR PUBLICATION IN IEEE TRANSACTIONS ON INTELLIGENT TRANSPORTATION SYSTEMS}%
{Shell \MakeLowercase{\textit{et al.}}: A Sample Article Using IEEEtran.cls for IEEE Journals}
%

\maketitle

\begin{abstract}
Existing driving style recognition systems largely depend on low-level sensor-derived features for training, neglecting the rich semantic reasoning capability inherent to human experts. This discrepancy results in a fundamental misalignment between algorithmic classifications and expert judgments. To bridge this gap, we propose a novel framework that integrates Semantic Privileged Information (SPI) derived from large language models (LLMs) to align recognition outcomes with human-interpretable reasoning. First, we introduce DriBehavGPT, an interactive LLM-based module that generates natural-language descriptions of driving behaviors. These descriptions are then encoded into machine learning-compatible representations via text embedding and dimensionality reduction. Finally, we incorporate them as privileged information into Support Vector Machine Plus (SVM+) for training, enabling the model to approximate human-like interpretation patterns.  Experiments across diverse real-world driving scenarios demonstrate that our SPI-enhanced framework consistently outperforms a wide range of baselines,  achieving $F_1$-scores of $90.2\%$ (car-following) and $91.0\%$ (lane-changing). Importantly, SPI is exclusively used during training, while inference relies solely on sensor data, ensuring computational efficiency without sacrificing performance. These results highlight the pivotal role of semantic behavioral representations in improving recognition accuracy while advancing interpretable, human-centric driving systems.
\end{abstract}

%
\IEEEpeerreviewmaketitle

\section{Introduction}

\IEEEPARstart{R}{cognizing} driving styles plays a pivotal role in understanding human-vehicle interactions, thereby improving personalized driving experience and enhancing the acceptance of advanced driver assistance systems \cite{liu2023real}. For example, adaptive cruise control systems offer configurable parameters, such as inter-vehicle distance, target speed, and driving modes, to accommodate both aggressive drivers prioritizing traffic throughput efficiency and conservative drivers emphasizing safety \cite{kim2023satisfactory,hang2022conflict}. Previous research has demonstrated the applicability of driving style recognition in intelligent vehicle systems including longitudinal control (e.g., personalized adaptive cruise control \cite{gao2020personalized}), vertical dynamics (e.g., active suspension optimization \cite{wang2015study}), and lateral guidance (e.g., steering assistance \cite{wang2016human} with lane keeping \cite{wang2018learning}).

\begin{figure}[t]
    \centering 
    \includegraphics[width=0.9\linewidth]{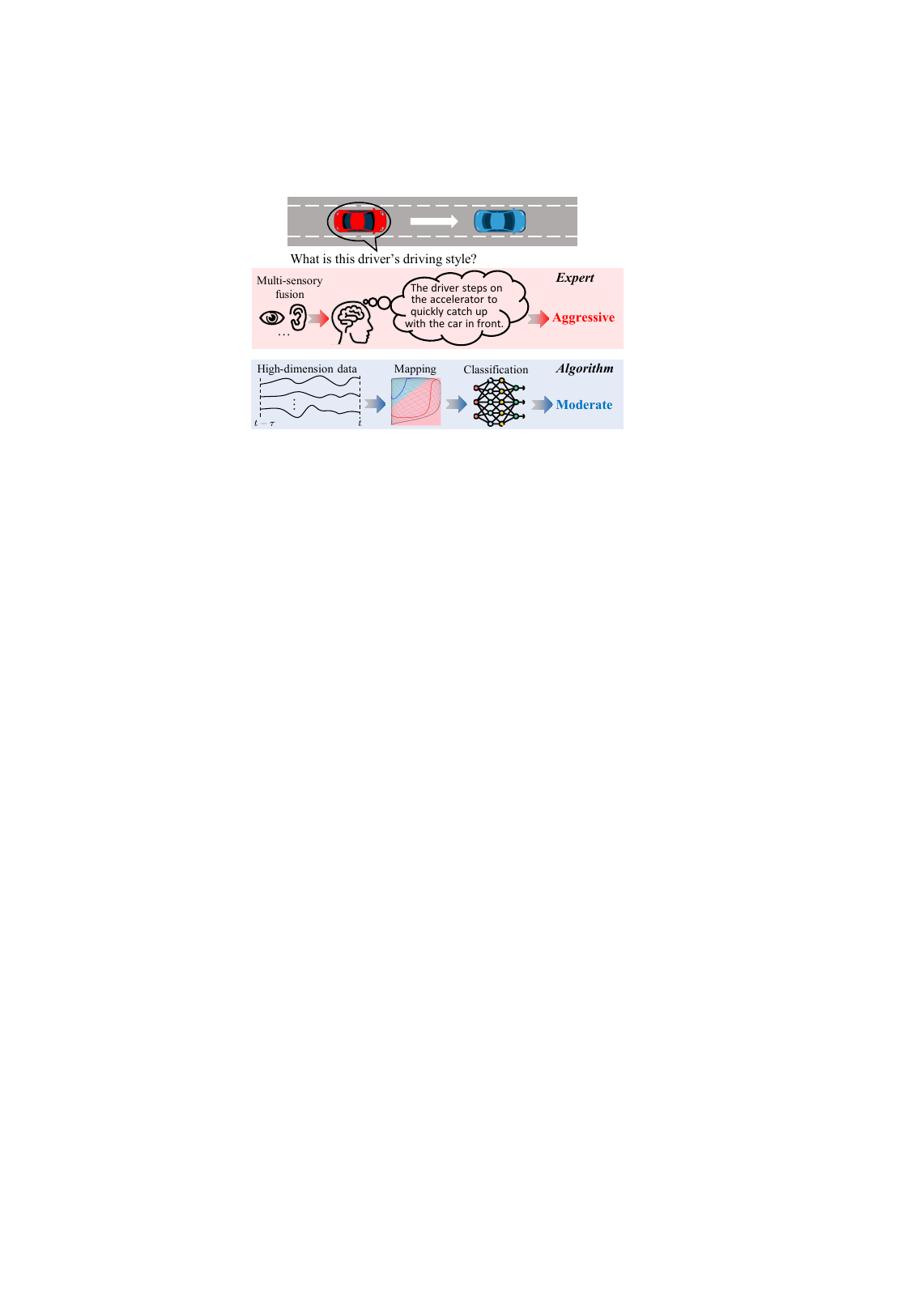}
    \caption{The difference between an expert and an algorithm for driving style recognition.}
    \label{fig: motivation}
\end{figure}

Existing approaches to driving style recognition, including rule-based \cite{dorr2014online,huang2020study,lin2022driving}, learning-based \cite{zhang2024shareable,zhang2023scene,tian2022incorporating,wang2022recognition}, and hybrid methods \cite{zhang2024embedded,chen2025knowledge}), primarily rely on low-level vehicular sensor data (e.g., speed, acceleration, steering wheel angle). While these methods capture physical driving behavior, they fail to emulate the human ability to infer high-level semantic cues, such as intent, environmental context, or social interactions (Fig. \ref{fig: motivation}) \cite{zhong2017theory, chen2025intention, chen2024fw, yang2024driving}. This semantic gap limits their capacity to model the underlying causality of driving behaviors, often resulting in misalignment between algorithmic outputs and human subjective evaluation. To bridge this gap, we argue that integrating semantically rich representations of driving behavior is essential --- not only to enhance recognition accuracy but also to improve the interpretability and human-consistency of machine-generated results.

While semantic information can be manually annotated by human experts, this approach suffers from inefficiency and poor scalability, particularly for large-scale naturalistic driving datasets. Consequently, a critical challenge is the development of an automated approach capable of extracting high-quality semantic representations of driving behavior. Recent advances in Large Language Models (LLMs) have revolutionized natural language processing, demonstrating exceptional capabilities in semantic comprehension and generative tasks \cite{min2023recent}. Some emerging studies have explored LLMs for autonomous driving agents \cite{cui2024drive} or scene annotation \cite{fu2024drive}, but they predominantly focus on trajectory generation or scenario analysis rather than real-time driving style recognition. More critically, they rely on computationally intensive online inference, creating a significant barrier for deployment on resource-constrained automotive electronic control units (ECUs).

To bridge this gap, we propose a novel framework that decouples semantic reasoning from real-time execution. We propose DriBehavGPT, an LLM-based module that integrates Chain-of-Thought reasoning \cite{wei2022chain} with structured prompt engineering to generate semantically rich driving behavior descriptions. The module processes raw in-vehicle sensor data and autonomously produces semantic behavior descriptions comparable to expert human annotations. Crucially, to circumvent the deployment constraints of LLMs, we incorporate these descriptions into a Learning Using Semantic Privileged Information (LUSPI) strategy \cite{pechyony2010theory}. By treating the LLM-generated semantics as privileged information utilized exclusively during training, our approach allows the deployed model to operate solely on real-time CAN bus signals. This design reconciles high-level semantic interpretability with the computational efficiency required for online implementation. This is the \textit{first} framework to bridge high-level LLM semantics and low-level sensor data through privileged information distillation for real-time driving style recognition. The main contributions are summarized as follows:
\begin{enumerate}
    \item  We introduce a privileged information-driven learning framework that exploits SPI during training to enhance recognition accuracy while retaining real-time inference capability during deployment.
    \item We develop DriBehavGPT, an LLM-based module designed to generate expert-like semantic descriptions of driving behaviors, which are subsequently transformed by a feature quantification pipeline into compact privileged vectors for model training, effectively bridging the semantic gap between qualitative expert descriptions and quantitative kinematic signals.
    \item Extensive experiments demonstrate that our framework significantly outperforms data-driven baselines in both accuracy and robustness. The results validate the effectiveness of leveraging semantic knowledge in the learning process, providing a reliable foundation for human-centric intelligent driving.
\end{enumerate}

The rest of this paper is structured as follows. Section \ref{sec: Literature Review} analyzes prior work on driving style recognition using semantic information. Section \ref{sec: The LUSPI Framework} describes the LUSPI framework and theoretical underpinnings. Section \ref{sec:Data Collection and Driving Style Labeling} describes real-world experiments and data collection.  Section \ref{sec: Experimental Settings} demonstrates the experimental settings, followed by the results and analysis in Section \ref{sec: Experimental Verification and Result Analysis} and conclusions in Section \ref{sec: Conclusions}.
\section{Literature Review}
\label{sec: Literature Review}
\subsection{Driving Behavior Semantics}
Semantic cognition plays a fundamental role in interpreting complex driving behaviors, as evidenced by neuroscientific and computational studies. Semantic abstraction mechanisms analogous to linguistic understanding are activated during driving tasks \cite{xia2023understanding}. This insight has motivated computational approaches to encode driving behaviors into semantically meaningful patterns. Early studies relied on human-provided semantic labels (e.g., survey on overtaking \cite{gulian1989dimensions} or speeding \cite{french1993decision}), while recent advances employ unsupervised learning to derive interpretable representations. Wang et al. \cite{wang2018driving} decomposed car-following behaviors into 75 primitive semantic patterns using non-parametric Bayesian modeling, whereas Chen et al. \cite{chen2021exploring} identified five latent semantic topics for lane-changing behaviors using latent Dirichlet allocation (LDA). Further refining this paradigm, Zhang et al. \cite{zhang2024shareable} introduced a hierarchical LDA to assign style-specific semantics to driving segments. Collectively, these works bridge low-level sensor data with high-level behavioral interpretations. However, these probabilistic models typically yield latent numerical topics rather than explicit linguistic descriptions. Recently, LLMs have demonstrated exceptional capabilities in generating human-like semantic reasoning, offering a more intuitive paradigm for encoding behavioral nuances.

\subsection{Learning Using Privileged Information (LUPI)}
The Learning Using Privileged Information (LUPI) paradigm, introduced by Vapnik and Vashist \cite{vapnik2009new}, extends conventional supervised learning by incorporating auxiliary knowledge (e.g., expert explanations or contextual metadata) during training. Unlike traditional frameworks where such information is discarded, LUPI leverages it to constrain the hypothesis space, often yielding superior generalization. The seminal SVM+ algorithm \cite{vapnik2009new} demonstrates this by modeling slack variables via privileged features. Subsequent work validated LUPI’s efficacy across domains: in computer vision, privileged descriptors improved image classification \cite{li2023multi,yang2017person,ortiz2023does}, disease diagnosis \cite{ganaie2022ensemble,gao2019privileged}, action recognition \cite{shi2017learning,liu2020exploring}. For instance, in image classification, incorporating privileged information (e.g., feature descriptions provided during training) can significantly improve classifier performance \cite{vapnik2009new,sabeti2020learning}.

\subsection{Driving Style Recognition Using Privileged Information}
Although LUPI remains underexplored in driving behavior analysis, its potential is underscored by studies integrating human prior knowledge.  Early rule-based systems codified expert-defined semantics (e.g., ``aggressive'' thresholds for acceleration), which lacked the flexibility to interpret complex driving behaviors. Recently, LLMs have emerged as a powerful tool for behavioral reasoning \cite{fu2024drive,cui2024drive}. For example, Fu et al. \cite{fu2024drive} demonstrated LLMs' ability to annotate driving scenarios with cognitively plausible semantics, though real-world validation is pending. Hybrid approaches, such as Zhang et al.'s \cite{zhang2024embedded} hierarchical fusion of expert rules and data-driven models, partially leverage privileged information but decouple it from the training loop. In contrast, LUPI embeds semantic guidance directly into the learning process, offering a principled framework to unify machine learning with domain knowledge --- a direction ripe for exploration in driving style recognition.


\begin{figure*}[t]
    \centering
\includegraphics[width=0.98\linewidth]{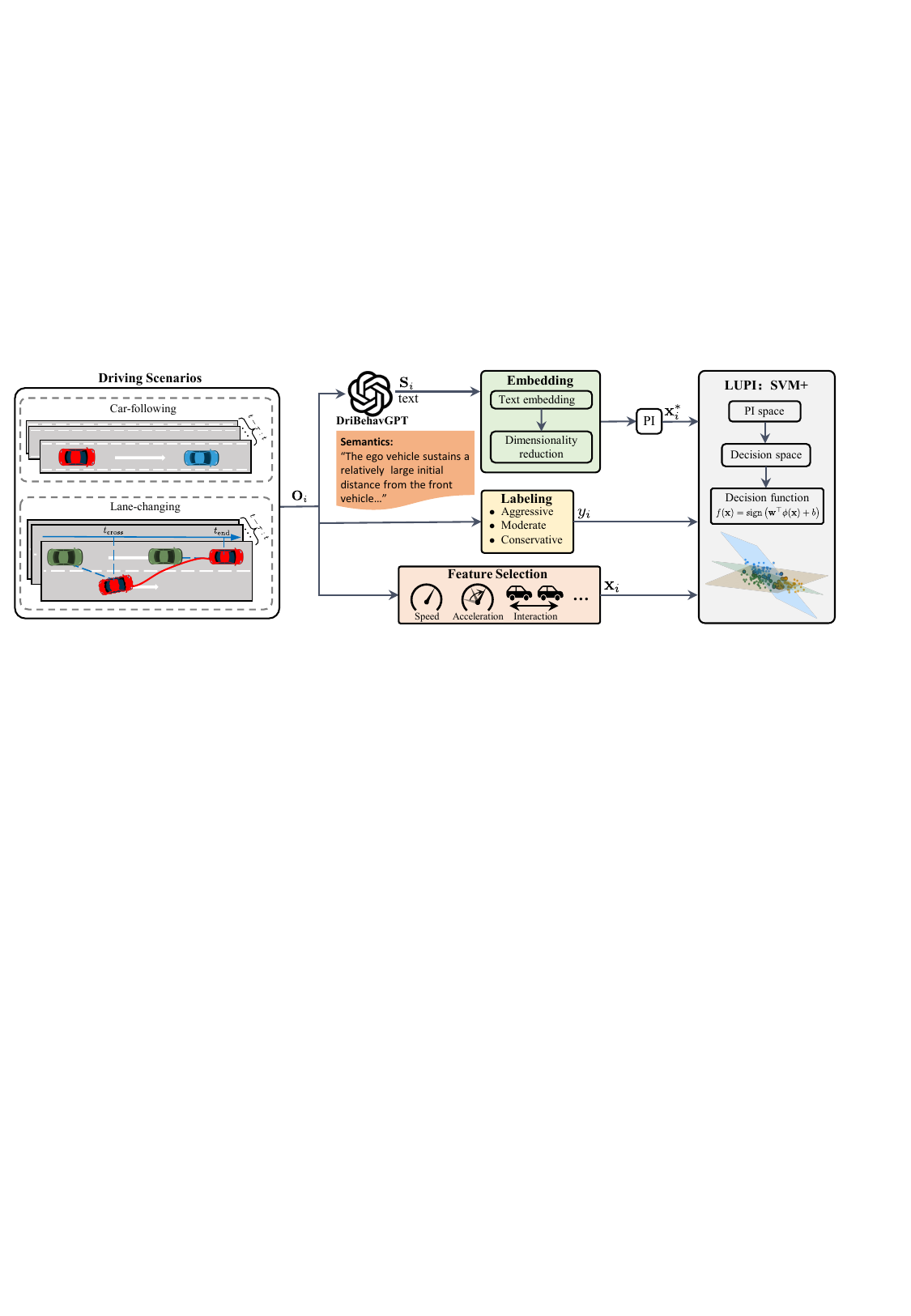}
    \caption{The LUSPI framework for driving style recognition. The input consists of the $i$-th sequential driving data $\mathbf{O}_i$ collected from diverse scenarios (i.e., car-following and lane-changing). These data are processed by DriBehavGPT to generate driving behavior semantic descriptions $S_i$, which are further transformed into privileged information (PI) $\mathbf{x}_i^*$ through text embedding and dimensionality reduction.}
    \label{fig: framework}
\end{figure*}

\section{The LUSPI Framework}
\label{sec: The LUSPI Framework}

The proposed LUSPI framework for driving style recognition (Fig. \ref{fig: framework}) comprises three core modules:  DriBehavGPT, Embedding, and LUPI. The LUPI module takes driving behavior features, style labels (Section \ref{sec:Data Collection and Driving Style Labeling}), and semantic privileged information generated by the DriBehavGPT module. DriBehavGPT translates sequential driving behavior data into textual descriptions, which the Embedding module then converts into compact numerical representations. By integrating data-driven features with the semantic reasoning capabilities of LLM, LUSPI enhances both the accuracy and interpretability of driving style recognition.

\subsection{LUPI: Support Vector Machine Plus (SVM+)}

In the LUPI paradigm, training samples contain additional privileged information unavailable during testing. Formally, the training set is denoted as $\left\{\left(\mathbf{x}_i, \mathbf{x}^*_i, y_i\right)\right\}_{i=1}^{n}$, where $n$ is the number of driving segments, $\mathbf{x}_i \in \mathbb{R}^D$ is the $D$-dimensional feature vector of the $i$-th segment, $\mathbf{x}_i^* \in \mathbb{R}^{D^*}$ is its associated $D^*$-dimensional privileged semantic information describing driving behavior, and  $y_i\in \mathcal{Y} = \{\mathrm{aggressive, moderate, conservative}\}$ is the driving style label. Since standard SVM+ is a binary classifier, we extend it to multi-class classification using a one-vs.-one strategy \cite{li2020stock}. For our three-class style recognition, this involves constructing pairwise classifiers where privileged information refines the specific boundary between each pair of styles and determining final predictions through majority voting. For clarity, we present the formulation for binary classification. The goal is to learn a decision function $f(\mathbf{x}):\mathbb{R}^{D}\to\mathcal{Y}$ mapping inputs to driving style labels. 

Similar to SVM, Support Vector Machine Plus (SVM+)\cite{vapnik2009new} defines the decision function as:
\begin{equation}
\label{eq: svm}
f(\mathbf{x}) = \text{sign}\left(\mathbf{w}^\top \phi(\mathbf{x}) + b\right)
\end{equation}
where $\mathbf{x}=\{\mathbf{x}_i\}_{i=1}^{n}$, $\mathbf{w}$ is the weight vector, $b$ is the bias term, $\phi(\cdot)$ is a feature mapping that enhances linear separability. We obtain the decision function by optimizing $\mathbf{w}$ and $b$. The standard SVM optimization problem is:
\begin{equation}
\begin{aligned}
  \min_{\mathbf{w},b,\xi} \quad & {\frac{1}{2}\|\mathbf{w}\|^2} + {C\sum_{i=1}^{n}\xi_i} \\
\text{s.t.} \quad & y_i(\mathbf{w}^\top \phi(x_i) + b) \geq 1 - \xi_i, \\
& \xi_i \geq 0, \quad \forall i = 1,\dots,n
\end{aligned}
\end{equation}
where $\xi_i$ is the slack variable for classification errors, and $C$ controls the trade-off between margin width and classification error. 

The SVM+ incorporates privileged information to refine the slack variables, leading to the following optimization:

\begin{equation}\label{eq: svm+ problem}
\begin{aligned}
    \min _{{\mathbf{w}^*}, b^*, \mathbf{w}, b} 
    \quad& {\frac{1}{2}\|\mathbf{w}\|^2} + 
          \textcolor{blue}{{\frac{\gamma}{2}\|{\mathbf{w}^*}\|^2}} + 
          \textcolor{red}{{C \sum_{i=1}^n \xi\left({\mathbf{w}^*}, b^*, \psi\left({\mathbf{x}_i^*}\right)\right)}}\\  
    \text { s.t. } 
    \quad& y_i\left(\mathbf{w}^{\top} \phi\left(\mathbf{x}_i\right)+b\right) \geq 1-\xi\left({\mathbf{w}^*}, b^*, \psi\left({\mathbf{x}_i^*}\right)\right), \\ 
    & \xi\left({\mathbf{w}^*}, b^*, \psi\left({\mathbf{x}_i^*}\right)\right) \geq 0, \quad i=1,\dots,n \\
\end{aligned}
\end{equation}
where $\textcolor{red}{\xi\left({\mathbf{w}^*}, b^*, \psi\left({\mathbf{x}_i^*}\right)\right)={\mathbf{w}^*}^{\top} \psi({\mathbf{x}_i^*})+b^*}$ is a slack function defined in the privileged semantic information space, $\psi(\cdot)$ is a feature mapping for privileged data, and ${\mathbf{w}^*}$ and $b^*$ are the weight vector and bias term,  respectively. SVM+ introduces the following improvements in the model architecture: (i) SVM+ replaces the slack variable $\xi_i$ with $\textcolor{red}{\xi\left({\mathbf{w}^*}, b^*, \psi\left({\mathbf{x}_i^*}\right)\right)}$ to adjust classification error tolerance based on driving semantic privileged information, improving consistency with human subjective evaluations. (ii) SVM+ simultaneously optimizes both the driving features (governed by $\mathbf{w}$) and the semantic privileged data, regularized by $\textcolor{blue}{\frac{\gamma}{2}\|{\mathbf{w}^*}\|^2}$, with a hyperparameter $\textcolor{blue}{\gamma}$ controlling the strength of semantic information on decision-making. The integration mechanism leverages semantic privileged information to adaptively refine the slack variables within the SVM+ objective function. While the primary weight vector $\textbf{w}$
 is optimized based on kinematic signals, the privileged features $\textbf{x}_i^*$ provide a supervisory adjustment for error modeling. This ensures that training samples characterized by significant kinematic variance but consistent semantic intent are optimally regularized, thereby enhancing the model’s generalization and robustness.
This embeds human-like semantic descriptions into the geometric structure of the model and generates dynamically adaptive driving style classification boundaries.

To solve (\ref{eq: svm+ problem}), we construct the Lagrangian function (Appendix A) and derive the dual form of SVM+ as:

\begin{equation}\label{eq:SVM+}
\begin{aligned}
& \max_{\boldsymbol{\alpha}, \boldsymbol{\beta}} 
\sum_{i=1}^{n} \alpha_i - \frac{1}{2} \sum_{i,j=1}^{n} \alpha_i \alpha_j y_i y_j K(\mathbf{x}_i, \mathbf{x}_j) \\
& \quad \quad - \frac{1}{2\gamma} \sum_{i,j=1}^{n} (\alpha_i + \beta_i - C)(\alpha_j + \beta_j - C) K^*(\mathbf{x}_i^*, \mathbf{x}_j^*) \\
&\hspace{1.5em} \text{s.t.} \quad \sum_{i=1}^{n} (\alpha_i + \beta_i - C) = 0, \\
&\hspace{3.7em} \sum_{i=1}^{n} y_i \alpha_i = 0, \\
&\hspace{4em} \alpha_i \geq 0, \quad \beta_i \geq 0.
\end{aligned}
\end{equation}
where $K(\mathbf{x}_i, \mathbf{x}_j) = \phi(\mathbf{x}_i)^\top \phi(\mathbf{x}_j)$ is a kernel function for driving features, and $K^*(\mathbf{x}^*_i, \mathbf{x}^*_j) = \psi(\mathbf{x}^*_i)^\top \psi(\mathbf{x}^*_j)$ represents a kernel function for semantic privileged information. We adopt the widely-used Gaussian kernel function\cite{xiao2014parameter}. To address the heterogeneity between driving features $x$ and privileged semantic
information $x^*$, we employ a decoupled kernel strategy, such as $K(\mathbf{x}_i, \mathbf{x}_j) = \exp(-\lambda \|\mathbf{x}_i - \mathbf{x}_j\|^2)$, $K(\mathbf{x}_i^*, \mathbf{x}_j^*) = \exp(-\lambda^* \|\mathbf{x}_i^* - \mathbf{x}_j^*\|^2)$, where distinct parameters $\lambda$ and $\lambda^*$ are optimized to capture the specific manifold structures of the sensor data and the text embeddings respectively.

We solve the optimization problem using the Sequential Minimal Optimization (SMO) algorithm \cite{pechyony2010smo} (Appendix B). Unlike standard SVM, the optimization of Lagrange multipliers $\alpha$ and $\beta$  in Eq.(\ref{eq:SVM+})  is coupled through the constraint $\sum(\alpha_i + \beta_i -C) = 0$. Therefore, we implement a specialized solver that iteratively updates $(\alpha, \beta)$ pairs to satisfy these specific constraints derived from the privileged information framework\cite{vapnik2009new}.
From the optimal Lagrange multipliers and the stationary conditions of the Lagrangian ($\nabla_{\mathbf{w}} L = 0$, $\nabla_{\mathbf{w}^*} L = 0$), the weight vectors $\mathbf{w}$ and $\mathbf{w}^*$ are derived as

\begin{equation}
\label{eq: w}
\begin{aligned}
    & \mathbf{w}=\sum_{i=1}^n \alpha_i y_i \phi\left(\mathbf{x}_i\right), \\
    & {\mathbf{w}^*}=\frac{1}{\gamma} \sum_{i=1}^n\left(\alpha_i+\beta_i-C\right) \psi\left(\mathbf{x}^*_i\right)
\end{aligned}
\end{equation}

Substituting (\ref{eq: w}) into (\ref{eq: svm}) yields the decision function, which depends only on the driving features' weight vector.
\begin{equation}
    f(\mathbf{x}) = \text{sign}\left( \sum_{i=1}^{n} \alpha_i y_i K(\mathbf{x}_i, \mathbf{x}) + b \right)
\end{equation}
where the bias term $b$ computed using all support vectors
\begin{equation}
b = \frac{1}{N_{SV}} \sum_{i \in SV} \biggl( y_{i} - \sum_{j=1}^{n} \alpha_{j} y_{j} K(\mathbf{x}_{j}, \mathbf{x}_{i}) \biggr)
\end{equation}
Here, $SV=\{i\mid\alpha_i>0\}$ denotes the set of support vector indices (i.e., driving segments influencing the decision hyperplane), and $N_{SV}$ is their cardinality. A key implementation advantage of this module is the decoupling of offline training and online inference. While LLM-generated semantic privileged information effectively reshapes the decision boundary during training, it is not required for real-time deployment. The final decision function $f(x)$ relies exclusively on the primary weight vector $w$ and sensor data, thereby eliminating the computational latency of LLMs and ensuring real-time execution on automotive ECUs.

\subsection{DriBehavGPT}
\begin{figure}
    \centering
    \includegraphics[width=0.85\linewidth]{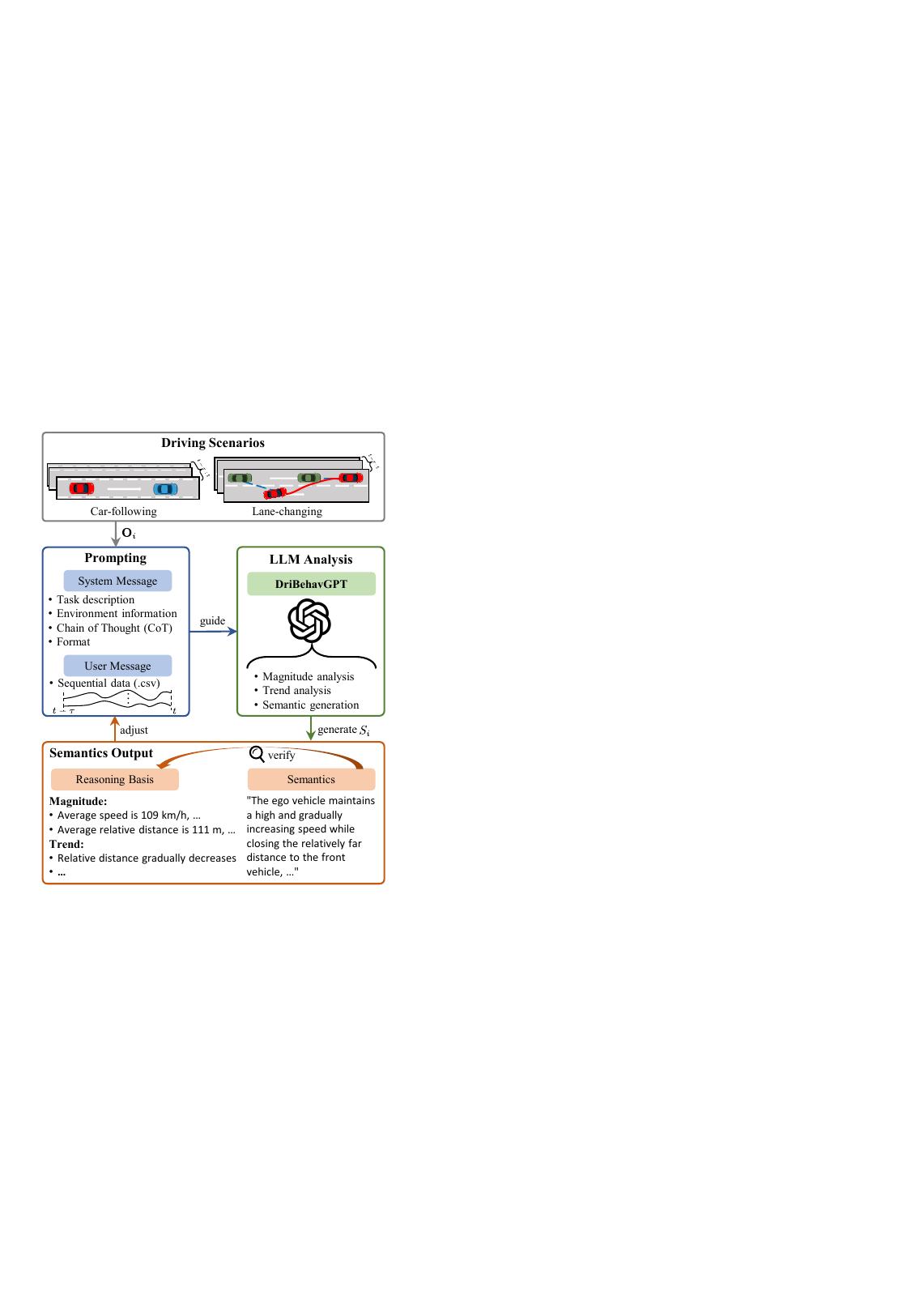}
    \caption{DriBehavGPT for semantically describing driving behaviors.}
    \label{fig: DriveGPT}
\end{figure}

We propose DriBehavGPT (Fig.~\ref{fig: DriveGPT}), an LLM-based module to generate semantics of driving behavior. The task is formulated as a language modeling problem. Structured prompts, designed through prompt engineering, guide DriBehavGPT in translating the $i$-th sequential driving observations $\mathbf{O}_i$ (e.g., speed, acceleration, relative distance/speed) into human-interpretable semantic description $S_i$ (e.g., `\textsf{\footnotesize The ego vehicle maintains a high and slightly increasing speed...}'). DriBehavGPT operates under a zero-shot learning paradigm, eliminating task-specific training by leveraging pretrained LLMs such as ChatGPT-4 \cite{achiam2023gpt}. To ensure transparency and reliability, we integrate a Chain-of-Thought mechanism into the prompt design, enabling the model to generate semantic description alongside their analytical reasoning. This allows systematic verification and refinement of the semantic description, ensuring their alignment with the derived reasoning basis. By jointly producing behaviors and their justification, DriBehavGPT improves interpretability, traceability, and alignment with domain knowledge while preserving the LLM's contextual understanding.

\subsubsection{Prompt Engineering}
\label{sec: prompt design}
To effectively leverage the reasoning and abstraction capabilities of large language models, we design structured prompts for DriBehavGPT that explicitly decompose driving behavior analysis into cognitively meaningful stages. Rather than treating the LLM as an opaque text generator, the prompt is constructed to align the model’s reasoning process with how human experts interpret driving behavior, thereby ensuring semantic reliability and task relevance. The input prompt comprises two components: a system message, which frames the LLM’s role and reasoning process, and a user message, which supplies driving data and contextual details. This dual-prompt design ensures that high-level semantic reasoning remains tightly coupled with observable vehicle dynamics, preventing unconstrained or speculative interpretations.

\textbf{System Message.}
This treats DriBehavGPT as an expert for driving behavior analysis, specifies task requirements, and employs Chain-of-Thought prompting to elicit stepwise reasoning. The message structures the analysis into three phases:
First, \textit{magnitude analysis}. Humans evaluate driving styles using perceptible quantitative features (e.g., speed and acceleration magnitude) \cite{elander1993behavioral}. Similarly, autonomous vehicle comfort is assessed via acceleration and jerk metrics \cite{bellem2018comfort}. To mirror human judgment, DriBehavGPT computes statistical features (e.g., mean speed, acceleration fluctuation range) from raw sequential driving data, forming an initial quantitative characterization of driving behavior. 
Second, \textit{trend analysis}. Human perception of driving styles relies on multisensory integration of temporal patterns \cite{lin2022effects}. Trend analysis supplements magnitude-based metrics by capturing dynamic contextual cues, enabling the LLM to distinguish between behaviors that are numerically similar but dynamically different which enhances holistic behavior interpretation. Finally, \textit{semantics generation}. Combining magnitude and trend features, DriBehavGPT utilizes the LLM's comprehension and generative capabilities to translate driving sequential data into expert-like natural language descriptions, encoding implicit factors such as aggressiveness, smoothness, and risk awareness.  The core principle of this design is contextual constraint decomposition, which prevents the LLM from making intuitive guesses. By enforcing a logical sequence from quantitative magnitude to temporal trend, the prompt ensures that the final semantic output is a rigorous deduction based on physical evidence, effectively emulating the cognitive process of experts.  

\textbf{User Message.}
The user message delivers formatted time-series driving data (i.e., CSV) with annotated columns to facilitate model comprehension.

\subsubsection{Interpretability of Driving Behavior Semantics}

To ensure transparency and credibility, this study integrates explicit explanatory requirements into the prompt design, guiding DriBehavGPT to furnish both semantic outputs and their analytical justifications. This approach elucidates the model's reasoning process, offering users actionable insight into its decision-making logic. Formally, the prompting-reasoning process can be defined as:
\begin{equation}
    \{\mathcal{S}, \mathcal{R}\}=F_{\mathrm{GPT}}(\mathcal{M}, \mathcal{U})
\end{equation}
where $F_{\mathrm{GPT}}$ denotes the DriBehavGPT module, it combines the system message $(\mathcal{M})$  and user message $(\mathcal{U})$, $\mathcal{S}$ is the semantic description of driving behaviors, and $\mathcal{R}$ provides the reasoning basis.

Unlike conventional methods that rely on opaque numerical features, DriBehavGPT frames driving behavior analysis as a language generation task. 
Different from raw sensor data that only captures what happened, these semantic features encapsulate latent behavioral priors, such as the driver's relative risk-aversion and tactical maneuvering intent which represent high-level abstractions of driving style that are otherwise invisible to purely data-driven models.
This paradigm shift enables human-understandable semantics (i.e., translating raw data into intuitive behavioral descriptors), traceable reasoning (i.e., exposing the causal linkages between data features and semantic conclusions), and dynamic interpretability (i.e., revealing how behavioral characteristics evolve temporally).

\subsection{Embedding}
\subsubsection{Text Embedding}
To process textual data with SVM+, we convert semantic descriptions into numerical representations using text embedding techniques. We employ SentenceTransformers \cite{reimers2019sentence} to generate contextualized embeddings that capture the semantics of driving behavior, ensuring compatibility with SVM+. The SentenceTransformers leverages transformer-based architectures to produce context-aware representations that capture rich intra- and inter-sentence relationships than traditional static word embeddings (e.g., Word2Vec\cite{mikolov2013efficient}, GloVe\cite{pennington2014glove}), yielding discriminative and semantically rich embeddings. This transformation allows high-level, human-interpretable behavioral semantics to be aligned with the numerical feature space required by the learning framework, while preserving relative semantic similarities between driving behaviors.

\subsubsection{Dimensionality Reduction}
The embeddings generated by SentenceTransformers typically have high dimensionality (several hundred dimensions), which may introduce computational inefficiency and feature redundancy. To address this, we implement Uniform Manifold Approximation and Projection (UMAP) \cite{healy2024uniform} for dimensionality reduction. Compared to alternative techniques (e.g., Principal Component Analysis and t-distributed Stochastic Neighbor Embedding), UMAP preserves both local and global structures in the data while efficiently handling nonlinear feature distributions. This ensures stable and compact representations of the semantic embeddings, which we then extract as privileged information $\mathbf{x}_i^*$ for SVM+.

This framework enhances interpretability by explicitly requiring the LLM to generate a systematic rationale $\mathcal{R}$ for each behavior description, ensuring that the source of privileged information is inherently transparent. By incorporating this rationale into the SVM+ paradigm, the training process aligns physical sensor observations with explicit semantic interpretations. Simultaneously, performance is improved as the framework incorporates latent contextual insights and behavioral intent into the optimization objective, which ensures that the resulting decision hyperplane is not solely dependent on fluctuating physical signals but is regularized by higher-level semantic constraints, leading to a more comprehensive and robust representation of driving styles.

\section{Data Collection and Driving Style Labeling}
\label{sec:Data Collection and Driving Style Labeling}
This section presents the methodology for data collection and driving style annotation in car-following and lane-changing scenarios. In existing driving behavior studies, car-following and lane-changing are typically treated as distinct behaviors and analyzed using separate modeling approaches\cite{chen2024review,mo2021physics,zheng2014recent}. Car-following focuses on continuous longitudinal interactions\cite{zhu2019typical,long2019research}, while lane-changing is commonly formulated as an event-driven lateral maneuver with different temporal and decision characteristics\cite{wang2021intelligent,zhang2023proactive}. Many established frameworks adopt a decoupled structure, where car-following serves as the fundamental behavior and lane-changing is introduced as a modular decision process\cite{yang2021personalized,zhang2018lane}. Such a decoupled formulation is also consistent with practical ADAS and autonomous driving architectures, where longitudinal car-following functions operate continuously, while lane-changing behaviors are activated in an event-driven manner by higher-level decision or planning modules and coordinated by exchanging only a limited set of high-level state information\cite{badue2021self}. In practical deployment, a scenario identification layer can be integrated into a hierarchical ADAS architecture which maintains car-following as the baseline state while dynamically triggering the lane-changing module via real-time signals such as lateral velocity exceeding $0.34$$\mathrm{m/s}$ or turn signal activation. This triggering mechanism enables the framework to switch seamlessly between specialized LUSPI models, ensuring that driving style recognition is always executed within the correct behavioral context. Following this widely adopted paradigm, we investigated driving style recognition in car-following and lane-changing scenarios separately. Data sources were selected based on specific hardware constraints: car-following analysis required the test vehicle's precise CAN-bus acceleration measurements to avoid differentiation noise. For lane-changing, the highD dataset \cite{krajewski2018highd} was necessary to capture rear-lag vehicle kinematics, as these were physically unobservable to the test vehicle's front-facing sensor. A unified annotation framework was applied to both datasets to ensure consistency in driving style labeling.

\begin{figure}[t]
    \centering
    \includegraphics[width=0.8\linewidth]{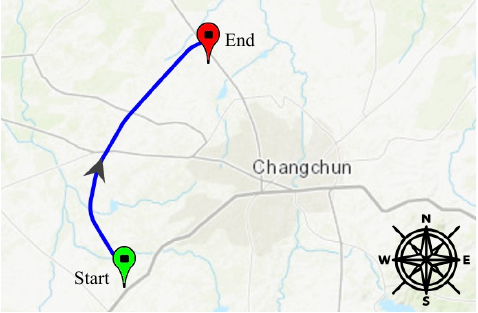}
    \caption{The driving routes of experiments in Changchun, China. }
    \label{fig: map_highways}
\end{figure}

\begin{figure}[t]
    \centering
    \includegraphics[width=\linewidth]{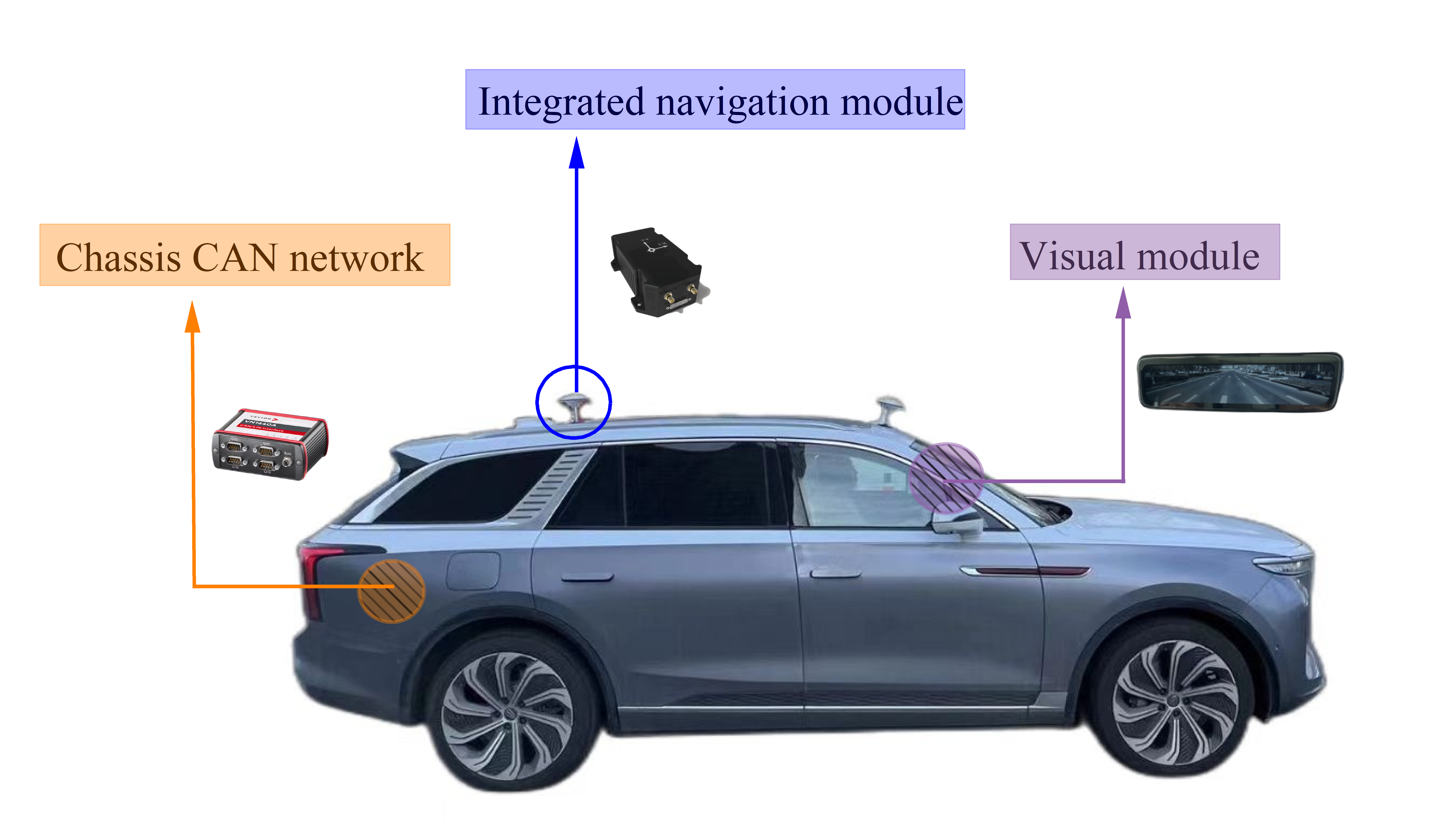}
    \caption{The testing vehicle equipped with a data-acquisition system.}
    \label{fig: vehicle}
\end{figure}

\subsection{Car-following Data}

\subsubsection{Driving Experiment}
To collect real human driving behavior data for method validation, we conducted large-scale naturalistic driving experiments involving $100$ participants across diverse demographics (occupation, gender, age: $21\sim 60$ years, and driving experience: $1\sim 35$ years). Detailed participant statistics are available in \cite{zhang2024100}. Experiments were performed on a $40.2$ km highway stretch (Fig.~\ref{fig: map_highways})  in Changchun, China. Participants operated the test vehicle (Fig.~\ref{fig: vehicle}) with a data acquisition system. The data acquisition system consists of three modules: the chassis CAN network for vehicle dynamics (speed, acceleration), an integrated navigation module for positioning, and a vision module for front-vehicle detection and relative distance estimation. This setup enabled recording of critical signals for car-following event extraction and driving style characterization. 

\subsubsection{Car-following Events Extraction}
To extract the car-following events, we selected four variables according to \cite{chen2023follownet,zhu2018human,wang2017capturing}:
Relative distance ($\Delta d_t^{\text{EV}}$) between the ego vehicle (EV) and front vehicle (FV), relative speed ($\Delta v_t^{\text{EV}}$) between the FV and the EV, ego vehicle speed ($v^{\text{EV}}_t$), ego vehicle longitudinal acceleration ($a^{\text{EV}}_{\text{lon},t}$). The car-following events were extracted from the collected naturalistic driving data based on three specific filtering criteria \cite{higgs2014segmentation}: (i) $\Delta d_t^{\text{EV}}\le120$ $\mathrm{m}$; (ii) $v^{\text{EV}}_t>16.7$ $\mathrm{m/s}$; (iii) The valid car-following state must be maintained continuously $\ge30$ $\mathrm{s}$. After filtering, $628$ car-following segments were retained for analysis.

\subsubsection{Feature Representation}
Driving styles were characterized via segment-wise mean values of kinematic features, ensuring computational efficiency and real-time use. The feature vector $\mathbf{x}_{i}^{\text{cf}} \in \mathbb{R}^{3}$ for the $i$-th car-following segment is:

\begin{equation}
    \begin{split}
    \mathbf{x}_{i}^{\text{cf}} = & \mathrm{mean} \{ v^{\text{EV}}_{\tau_i}, a^{\text{EV}}_{\text{lon},{\tau_i}}, \text{TTC}_{\tau_i}  \}^{\top}
    \end{split}
\end{equation}
where $\mathrm{mean}\{\cdot\}$ computes element-wise averages over the segment, $\text{TTC}_t=\Delta d^{\text{EV}}_t/\Delta v^{\text{EV}}_t$ is the time-to-collision (TTC) at time $t$, $\tau_i$ is the duration of the $i$-th segment. 

\subsection{Lane-changing Data}

\subsubsection{Definition of Lane-changing Events}
\label{sec:lane-change-definition}

\begin{figure}
    \centering
    \includegraphics[width=\linewidth]{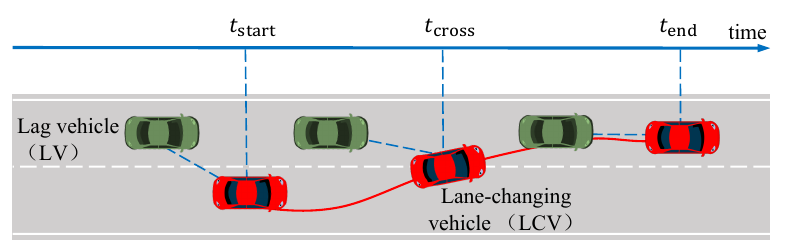}
    \caption{Illustration of key time points in the lane-changing behavior definition.}
    \label{fig: LC behavior definition}
\end{figure}

To automate lane-changing detection from large-scale trajectory data, we defined a lane-change maneuver as a complete lateral transition into an adjacent lane, segmented by three critical timestamps (Fig.~\ref{fig: LC behavior definition}) \cite{xing2020ensemble}:
(i) Start time ($t_\text{start}$) --- The instant when the vehicle's lateral velocity first exceeds $0.34$\,$\mathrm{m/s}$~\cite{zhang2020interaction}, marking the transition from lane-keeping to lane-changing initiation.
(ii) Crossing time ($t_\text{cross}$) --- The moment when the vehicle's centerline crosses the lane boundary.
(iii) End time ($t_\text{end}$) --- The time at which the lateral speed decays below $0.2$ $\mathrm{m/s}$, indicating maneuver completion.

\subsubsection{Lane-changing Events Extraction}

Following~\cite{xing2020ensemble, yan2023interaction}, we defined the lane-changing event extraction procedure based on behavioral completeness and interaction with a following vehicle. The extraction process comprises four steps. 
(i) \textbf{Key time points identification}: For each lane-changing event, we extract the timestamps $t_\text{start}$, $t_\text{cross}$, and $t_\text{end}$ (defined in Section~\ref{sec:lane-change-definition}) and record the full trajectory of the lane-changing vehicle over the interval $[t_\text{start}, t_\text{end}]$;
(ii) \textbf{Following vehicle assignment}: The lag vehicle is identified as the one spatially closest to the rear of the lane-changing vehicle at $t_\text{end}$ and its ID is stored for analysis;
(iii) \textbf{Event completeness validation}: Ensure that the lag vehicle remains within the recording range and stays in the same lane throughout $[t_\text{start}, t_\text{end}]$;
(iv) \textbf{Interaction validation}: To ensure meaningful interaction, only events where the time headway at $t_\text{cross}$ is less than two seconds are retained \cite{zhang2020interaction}. Applying this procedure to the highD dataset yields $633$ valid lane-changing events for further analysis.

\subsubsection{Lane-changing Feature Selection}

Based on prior research \cite{yang2019time, asaithambi2017overtaking}, we characterized the lane-changing style from two aspects: (i) the relative motion between vehicles and (ii) the lateral control of the lane-changing vehicles. We consider the following variables: relative distance ($\Delta d_{t}^{\text{LCV}}$) and relative speed ($\Delta v_{t}^{\text{LCV}}$) between the lane-changing vehicle (LCV) and the lag vehicle in the target lane, and LCV's lateral acceleration ($a_{\text{lat},t}^{\text{LCV}}$). These variables provide intuitive physical interpretations of driving styles: Conservative drivers maintain larger inter-vehicle gaps, respond more promptly, and exhibit smoother lateral movements, while aggressive drivers tend to adopt smaller gaps, exhibit delayed evasive actions, and apply more abrupt steering. We used the mean values of the above variables within the segment as the features. For the feature vector of the $i$-th lane-changing segment $\mathbf{x}_{i}^{\text{lc}}$, we represented it using the mean values of these variables over the segment duration  $\tau_i'$:

\begin{equation}
    \begin{split}
    \mathbf{x}_{i}^{\text{lc}} = & \mathrm{mean} \{ \Delta d_{\tau_i'}^{\text{LCV}}, \Delta v_{\tau_i'}^{\text{LCV}}, a_{\text{lat},\tau_i'}^{\text{LCV}}\}^{\top}.
    \end{split}
\end{equation}

\subsection{Driving Style Labeling}

\subsubsection{Car-following Style Labeling}
\label{sec: Car-following Data Labeling}

\begin{figure}
    \centering
    \includegraphics[width=0.95\linewidth]{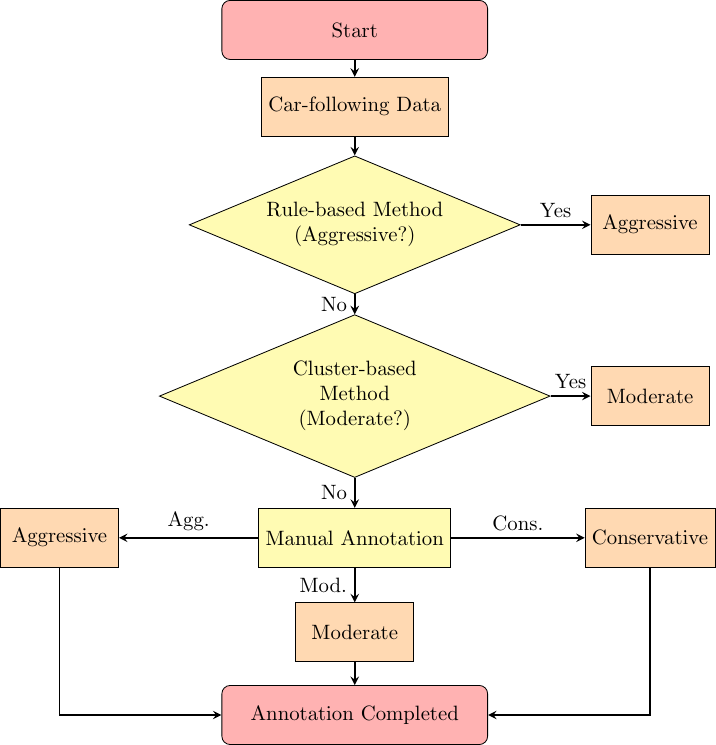}
    \caption{Three-stage annotation workflow for driving style labeling.}
    \label{fig: FlowLabeling}
\end{figure}

\begin{figure*}
    \centering
    \includegraphics[width=1\linewidth]{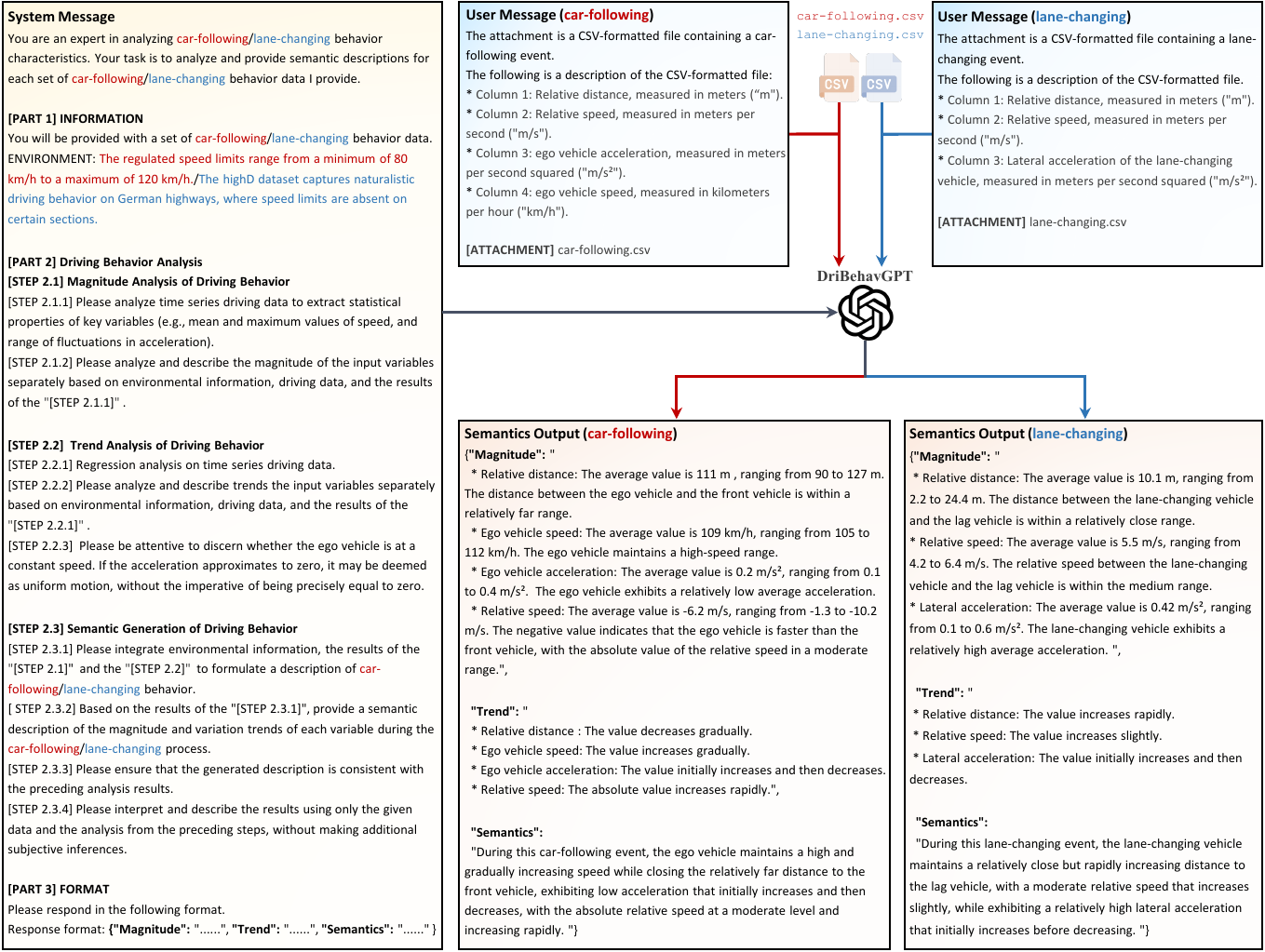}
    \caption{Prompt design for DriBehavGPT in car-following (\textcolor{red}{red mark}) and lane-changing (\textcolor{blue}{blue mark}) scenarios.}
    \label{fig: Prompt-CF}
\end{figure*}

Three primary methods exist for driving style annotation: manual labeling, rule-based classification, and clustering techniques. Manual annotation ensures high accuracy but is labor-intensive, while rule-based methods are computationally efficient but lack adaptability. Clustering techniques balance between automation and flexibility. To improve the reliability and interpretability of driving style annotations, we adopted a three-stage annotation pipeline (Fig.~\ref{fig: FlowLabeling}) instead of a single-step labeling strategy. This design is motivated by recent driving style recognition studies, which show that driving style understanding benefits from progressive reasoning and the incorporation of prior knowledge rather than direct end-to-end inference \cite{zhang2024embedded}. In practice, raw trajectory data often contain ambiguous or context-dependent behaviors, making one-shot annotation unreliable. Accordingly, each stage of the proposed pipeline is designed to address a specific source of annotation uncertainty and progressively refine the semantic interpretation of driving styles. The three-stage annotation pipeline we proposed processes car-following data sequentially through the following steps: 
(i) \textbf{Rule-based aggressive style detection:}  Aggressive driving styles are identified by thresholding two critical variables: speed and TTC. Samples exceeding $120$ km/h (the legal highway limit in the data collection region) or with TTC $<5$s \cite{vogel2003comparison} are labeled as aggressive. The motivation is to isolate extreme behaviors using explicit safety limits, which ensures that unambiguous high-risk maneuvers are identified with high
 consistency and zero manual effort. (ii) \textbf{Clustering-based moderate style identification:} Following the removal of extreme cases, samples not categorized as aggressive are subjected to DBSCAN clustering \cite{schubert2017dbscan} based on a feature set comprising vehicle speed, acceleration, and TTC to distinguish moderate driving behaviors. The underlying motivation for this approach is that moderate driving style typically constitutes the statistical majority in naturalistic datasets, forming a dense core within the kinematic feature space. By automating the identification of these standard patterns, this stage effectively streamlines the annotation process, thereby mitigating potential human fatigue and reducing the subjective bias inherent in manual labeling of typical driving behaviors. (iii) \textbf{Manual annotation:} The final stage involves four independent experts to label ambiguous transitional samples. The motivation is to utilize human expert intuition for cases where statistical boundaries overlap. Effectiveness is ensured through a majority-voting mechanism, which acts as a final quality-control layer to ensure high-fidelity labels for model training. The pipeline labeled $183$ aggressive, $307$ moderate, and $138$ conservative car-following samples.

\subsubsection{Lane-Changing Style Labeling}

The lane-changing data were annotated using the three-stage driving style labeling procedure (Section~\ref{sec: Car-following Data Labeling}). Based on statistical analysis of the lane-changing dataset and insights from~\cite{yang2019time}, we defined the following classification criteria. A lane-changing sample is labeled as aggressive if it satisfies at least one of the following conditions: Average relative distance to the preceding vehicle $<20$ $\mathrm{m}$, average relative speed $<0$ $\mathrm{m/s}$ (indicating closing-in behavior), or average lateral acceleration $>0.5$ $\mathrm{m/s^2}$ (reflecting abrupt steering maneuvers).  Samples not classified by the rule-based criteria were annotated using clustering and manual verification. The final labeled dataset comprised $216$ aggressive, $313$ moderate, and $104$ conservative lane-changing samples.

\subsection{Statistical Analysis of Input Data}

To quantitatively characterize the input data, we analyzed the statistical distribution of key kinematic features across the three driving styles. We calculated the mean and standard deviation (SD) for each feature to assess data dispersion. The detailed statistical results for both car-following and lane-changing scenarios are provided in Appendix C. The results show that while high intra-class variance exists—reflecting the diversity of individual driving habits—the distributions remain structurally distinct. Furthermore, we performed a one-way Analysis of Variance (ANOVA) to verify the statistical significance of the variations among different styles. The results confirm that the selected features exhibit statistically significant differences across the conservative, moderate, and aggressive categories (all $p$-values $< 0.001$). This ensures that despite the statistical variance of raw input signals, the identified styles possess clear boundaries for model training.

\section{Experimental Settings}
\label{sec: Experimental Settings}
We evaluate the proposed method in car-following and lane-changing scenarios. Below, we detail the prompt design for DriBehavGPT, the parameter configuration of the LUSPI module, and the evaluation metrics for both scenarios.

\subsection{Prompt Design for DriBehavGPT}
To extract car-following and lane-changing behavior semantics using large language models, we design scenario-specific prompts (Fig.~\ref{fig: Prompt-CF}).

\subsubsection{Car-following Scenario}

To accurately characterize dynamic car-following behaviors, DriBehavGPT integrates four dynamic features:
(i) Ego vehicle (EV) speed sequence ($v_{ \tau_{i}}^{\text{EV}}$), which reflects the driver's speed preference, (ii) Ego vehicle acceleration sequence   ($a_{\text{lon},\tau_{i}}^{\text{EV}}$), which captures the driver's acceleration habits, (iii) Relative distance sequence  ($\Delta d_{\tau_i}^{\text{EV}}$), which depicts the inter-vehicle gap dynamics, indicating risk tolerance, and (iv) Relative speed sequence ($\Delta v_{\tau_i}^{\text{EV}}$), which describes the relationship between the two vehicles. These features are transformed into semantic descriptions using a large language model. The input-output structure is

\begin{align*}
  \text{Input: }  & \mathbf{O}_{i}^{\text{cf}} = [v_{\tau_i}^{\text{EV}}, a_{\text{lon},\tau_i}^{\text{EV}}, \Delta d_{\tau_i}^{\text{EV}}, \Delta v_{\tau_i}^{\text{EV}}] \\
  \text{Output: } & S_{i}^{\text{cf}}
\end{align*}
where $S_i^{\text{cf}}$ is a text-based semantic description of the $i$-th car-following segment.

The user message provides the car-following time-series data (e.g., in CSV format, such as \texttt{car-following.csv}), with explicit column-wise descriptions (including units) to ensure precise interpretation. The system message follows the approach in Section~\ref{sec: prompt design}, encompassing environmental context, behavior magnitude/trend analysis, semantic generation guidelines, and output formatting rules. The combined system message, user message, and input data constitute the structured prompt, which is processed by DriBehavGPT to generate human-interpretable behavior semantics.

\subsubsection{Lane-changing Scenario}
DriBehavGPT captures dynamic lane-changing behaviors using three key features: (i) Relative distance sequence ($\Delta d_{\tau_i'}^{\text{LCV}}$), which captures the safety gap evolution between the lane-changing vehicle (LCV) and the lag vehicle (LV) in the target lane, (ii) Relative speed sequence ($\Delta v_{\tau_i'}^{\text{LCV}}$), which describes interaction dynamics between the LCV and the LV, and (iii) Lateral acceleration sequence  ($a_{\text{lat},\tau_i'}^{\text{LCV}}$), which represents the sharpness of steering maneuvers. These time-series features are processed by LLM to generate semantic descriptions. The model's input and output are defined as:
\begin{align*}
\text{Input} & : \mathbf{O}_i^{\text{lc}} = \left[\Delta d_{\tau_i'}^{\text{LCV}}, \Delta v_{\tau_i'}^{\text{LCV}}, a_{\text{lat},\tau_i'}^{\text{LCV}} \right] \\
\text{Output} & : S_i^{\text{lc}}
\end{align*}
where $S_i^{\text{lc}}$ is the text-based semantic description of the $i$-th lane-changing segment.

We employ a similar prompt structure for the lane-changing scenario. The user message provides specific lane-changing behavior data in CSV format (i.e., \texttt{lane-changing.csv}), complete with explicit column descriptions to maintain interpretability. Meanwhile, the system message retains the same analytical framework as described previously, allowing DriBehavGPT to generate corresponding behavior semantics for this different driving context.

\subsection{LUSPI Module Parameters}
We present the models and core parameters of the four modules in the LUSPI framework (Table \ref{tab: Parameters}). DriBehavGPT is constructed using OpenAI's GPT-4o API. To ensure the reproducibility and consistency of the generated semantics, DriBehavGPT is configured with a temperature parameter of 0, which promotes deterministic outputs by prioritizing the highest-probability tokens \cite{huang2025survey}. To ensure adaptability across different datasets and driving environments, we adopt a zero-shot prompting strategy rather than fine-tuning the LLM, thereby avoiding dataset-specific overfitting. DriBehavGPT is constrained by structured prompts that mandate a logical deduction from quantitative kinematic magnitudes to qualitative behavioral descriptions, thereby ensuring that the generated semantics are strictly grounded in objective physical evidence rather than unconstrained internal priors. Furthermore, the LLM functions exclusively as a source of privileged information during the training phase. This architectural decoupling ensures that any latent linguistic biases inherent in the pre-trained model are confined to the offline training objective, which prevents such biases from propagating to online inference that relies solely on objective in-vehicle sensor signals. Text embeddings are generated using the `all-mpnet-base-v2' model from SentenceTransformers \cite{reimers2019sentence}, which encodes semantic descriptions of car-following and lane-changing events into vectorized representations. Subsequently, we employ UMAP \cite{healy2024uniform} for dimensionality reduction and SVM+ for classification. To ensure optimal performance, the hyperparameters for both modules are determined via systematic grid search: the number of UMAP components is set to 5, while the SVM+ parameters are configured as $C=8, \gamma=1.1$ for the car-following scenario and $C=13, \gamma=1.3$ for the lane-changing scenario.

\begin{table}[t]
\centering
\setlength{\tabcolsep}{4pt} 
    \caption{\textcolor{black}{Models and Parameters of Modules in the LUSPI Framework}}\label{tab: Parameters}
\begin{tabular}{lll@{}}
\hline\hline
LUSPI Module              & Model                & Parameters         \\
\hline
 DriBehavGPT          & OpenAI GPT-4o        & temperature = 0                  \\
Text Embedding           & SentenceTransformers & all-mpnet-base-v2 \\
Dimensionality Reduction & UMAP                 & n\_components$=5$ \\
\multirow{2}{*}{LUPI}     & \multirow{2}{*}{SVM+} & cf: $C=8$, $\gamma=1.1$ \\ 
                         &                      & lc: $C=13$, $\gamma=1.3$ \\
\hline\hline
\end{tabular}
\end{table}

\begin{table*}[t]
\setlength{\tabcolsep}{4pt} 
\small 
\centering
\caption{\textcolor{black}{Performance Comparison ($F_1$-score \%) with Baseline Methods on Two Driving Scenarios. All Results are Reported as Mean ± SD (The Best Performance is Highlighted in Bold Red, and the Second Best Performance is Indicated by Underlined Blue)}}
\label{tab:Performance-Combined}
\renewcommand{\arraystretch}{1.4} 

\begin{tabular}{l|cccc|cccc} 
\hline\hline
\multirow{3}{*}{Method} & \multicolumn{4}{c|}{Car-following Scenario} & \multicolumn{4}{c}{Lane-changing Scenario} \\ \cline{2-9} 
 & Conservative & Moderate & Aggressive & Average & Conservative & Moderate & Aggressive & Average \\ \hline

KNN 
 & $80.2 \pm 3.5$ & $79.5 \pm 1.2$ & $69.8 \pm 2.4$ & $76.5 \pm 2.1$ 
 & $78.5 \pm 4.8$ & $82.4 \pm 1.1$ & $73.4 \pm 1.9$ & $78.1 \pm 2.4$ \\

SVM
 & $85.7 \pm 2.8$ & $85.8 \pm 0.9$ & $79.9 \pm 1.8$ & $83.8 \pm 1.6$ 
 & $80.6 \pm 3.9$ & $85.7 \pm 0.8$ & $86.8 \pm 1.5$ & $84.4 \pm 1.9$ \\

Random Forest
 & $85.5 \pm 2.5$ & $82.8 \pm 0.8$ & $83.7 \pm 1.6$ & $84.0 \pm 1.4$ 
 & $82.5 \pm 3.5$ & $87.2 \pm 0.7$ & $84.1 \pm 1.4$ & $84.6 \pm 1.7$ \\

LightGBM 
 & \textcolor{blue}{\underline{$87.5 \pm 2.2$}} & $87.0 \pm 0.7$ & $81.2 \pm 1.5$ & $85.2 \pm 1.2$ 
 & $85.2 \pm 3.1$ & $88.5 \pm 0.6$ & $84.1 \pm 1.3$ & $85.9 \pm 1.5$ \\

XGBoost
 & $86.0 \pm 2.3$ & $86.2 \pm 0.8$ & $81.6 \pm 1.5$ & $84.6 \pm 1.3$ 
 & $83.7 \pm 3.2$ & $88.0 \pm 0.6$ & $84.5 \pm 1.4$ & $85.4 \pm 1.6$ \\

CatBoost
 & $86.8 \pm 2.1$ & $87.0 \pm 0.7$ & $83.9 \pm 1.4$ & $85.9 \pm 1.1$ 
 & $86.0 \pm 3.0$ & \textcolor{blue}{\underline{$89.2 \pm 0.5$}} & \textcolor{blue}{\underline{$87.3 \pm 1.2$}} & \textcolor{blue}{\underline{$87.5 \pm 1.4$}} \\

MLP 
 & $86.5 \pm 2.9$ & \textcolor{blue}{\underline{$87.4 \pm 1.0$}} & $84.1 \pm 1.9$ & $86.0 \pm 1.7$ 
 & $84.9 \pm 3.8$ & $88.2 \pm 0.9$ & $84.8 \pm 1.6$ & $86.0 \pm 1.9$ \\

LLM-based 
 & $87.2 \pm 2.2$ & $86.3 \pm 1.1$ & \textcolor{red}{\textbf{88.5 $\pm$ 1.2}} & \textcolor{blue}{\underline{$87.3 \pm 1.5$}} 
 & \textcolor{blue}{\underline{$86.8 \pm 3.2$}} & $87.1 \pm 1.1$ & $85.9 \pm 1.8$ & $86.6 \pm 2.0$ \\ \hline 

\textbf{LUSPI (Ours)} 
 & \textcolor{red}{\textbf{92.7$^{*} \pm$ 2.1}} & \textcolor{red}{\textbf{90.7$^{*} \pm$ 0.6}} & \textcolor{blue}{\underline{$87.2 \pm 1.4$}} & \textcolor{red}{\textbf{90.2$^{*} \pm$ 1.4}} 
 & \textcolor{red}{\textbf{89.5$^{*} \pm$ 2.6}} & \textcolor{red}{\textbf{92.1$^{*} \pm$ 0.5}} & \textcolor{red}{\textbf{91.5$^{*} \pm$ 1.0}} & \textcolor{red}{\textbf{91.0$^{*} \pm$ 0.8}} \\ 

\hline\hline
\end{tabular}

\begin{minipage}{0.96\textwidth} 
    \vspace{1mm}
    \footnotesize
    \textbf{Note:} 
    The symbol ``\textcolor{red}{*}'' indicates that the performance improvement of the best method over all other compared methods is statistically significant ($p < 0.05$) based on paired $t$-tests across 5-fold cross-validation.
\end{minipage}

\renewcommand{\arraystretch}{1.0}
\end{table*}

\subsection{Baseline Comparison}

To validate the performance of our proposed framework in driving style recognition, we compare it with several machine learning algorithms: K-Nearest Neighbors (KNN) \cite{zhang2023scene}, SVM \cite{zhang2024embedded}, and Random Forest \cite{zhang2024100}, advanced ensemble learning frameworks \cite{zhang2024analysis}: LightGBM, XGBoost, and CatBoost, as well as neural network and semantic reasoning models: Multilayer Perceptron (MLP) \cite{chen2024sts} and LLM-based \cite{cui2024drive}. The details of each model are described below:

\textbf{KNN} acts as a non-parametric instance-based benchmark, determining driving styles based on the majority vote of the nearest neighbors within the kinematic feature space. \textbf{SVM} provides a foundational baseline to isolate the impact of privileged information; it employs a Gaussian kernel to classify styles using solely sensor data. \textbf{Random Forest} represents the bagging ensemble approach, constructing a multitude of decision trees during training to mitigate overfitting and enhance classification stability. \textbf{XGBoost} is selected for its scalable gradient boosting framework, which utilizes a level-wise tree growth strategy and robust regularization functions to effectively capture complex non-linear patterns. Unlike XGBoost, \textbf{LightGBM} adopts a leaf-wise tree growth strategy with depth limitations, a design that significantly optimizes training speed and memory efficiency for sequential data. \textbf{CatBoost} is distinguished by its implementation of symmetric binary trees and ordered boosting, which minimizes prediction shift and gradient bias to ensure robust performance. \textbf{MLP} represents a type of neural network approach; we design a standard feedforward neural network with fully connected layers to model the relationships between vehicular sensor inputs and driving styles. \textbf{LLM-based} serves as a semantic reasoning benchmark; instead of training, this method directly leverages the inferential capabilities of Large Language Models to analyze the vehicular sensor inputs and classify styles.

To ensure optimal performance, we apply systematic grid search to fine-tune the hyperparameters of all conventional algorithms. The final optimal parameter settings are detailed in Appendix D. For the LLM-based approach, we focus on rigorous prompt engineering to maximize inference accuracy. This guarantees a fair comparison by evaluating each method at its best capability.

\subsection{Evaluation Metrics}
We evaluate the performance of the comparative baselines using a 5-fold cross-validation protocol, prioritizing the $F_1$-score to effectively balance the trade-off between false positives and negatives in this multi-class task. To formulate these metrics for each driving style category, we use an integer iterator $z$, where $z=1$ corresponds to conservative, $z=2$ to moderate, and $z=3$ to aggressive. We define \( n_{\text{true}}(z) \) as the number of ground-truth samples of the $z$-th category, \( n_{\text{pred}}(z) \) as the number of samples classified as the $z$-th category, and \( n_{\text{correct}}(z) \) as the number of correctly predicted samples for the $z$-th category. The metrics are defined as follows.
\begin{itemize}
    \item Precision ($\eta_{\mathrm{pre},z}$) measures the model's reliability in minimizing false positives --- i.e., misclassifying other styles as the $z$-th category,
    \begin{equation}
        \eta_{\mathrm{pre},z} = \frac{n_{\text{correct}}(z)}{n_{\text{pred}}(z)}.
        \label{eq:precision}
    \end{equation}

    \item Recall ($\eta_{\mathrm{rec},z}$) quantifies the model's ability to avoid false negatives by capturing most instances of the $z$-th category, 
\begin{equation}
    \eta_{\mathrm{rec},z}= \frac{n_{\text{correct}}(z)}{n_{\text{true}}(z)}.
    \label{eq:recall}
\end{equation}
\item $F_1$-score ($F_{1,z}$) denotes the harmonic mean of precision and recall for the $z$-th category. We report this metric for each specific driving style individually, as well as the overall macro-average,

\begin{equation}
    F_{1,z} = \frac{2\eta_{\mathrm{pre},z} \eta_{\mathrm{rec},z}}{\eta_{\mathrm{pre},z} + \eta_{\mathrm{rec},z}}, \quad F_{1,\text{avg}} = \frac{1}{3} \sum_{z=1}^{3} F_{1,z}.
    \label{eq:f1_score}
\end{equation}

\end{itemize}

All results are reported as the mean ± standard deviation (mean ± SD) across the five folds, ensuring the statistical reliability and stability of the evaluations. Furthermore, to validate the significance of the performance improvements, we conduct paired $t$-tests between the proposed framework and the baseline methods. Improvements are considered statistically significant if the $p$-value is less than 0.05 ($p < 0.05$).

\section{Experimental Verification and Result Analysis}
\label{sec: Experimental Verification and Result Analysis}

\subsection{Overall Performance}
As summarized in Table \ref{tab:Performance-Combined}, the proposed LUSPI framework demonstrates the highest overall efficacy across both car-following and lane-changing scenarios, significantly outperforming baseline methods in terms of average evaluation metrics. In the car-following scenario, LUSPI achieved an average $F_1$-score of $90.2\% \pm 1.4\%$, representing a substantial margin over the traditional SVM baseline ($83.8\% \pm 1.6\%$) and the suboptimal LLM-based approach ($87.3\% \pm 1.5\%$). Specifically, LUSPI yields a $7.6\%$ improvement in the average $F_1$-score relative to SVM. In terms of fine-grained classification, LUSPI exhibits exceptional precision in identifying ``Conservative'' and ``Moderate'' behaviors, attaining the highest $F_1$-scores of $92.7\%$ and $90.7\%$, respectively. Regarding ``Aggressive'' behaviors, while the LLM-based benchmark marginally surpassed LUSPI ($88.5\%$ vs. $87.2\%$), LUSPI maintained a robust second-best performance. Notably, it outperformed all other supervised learning algorithms (e.g., CatBoost at $83.9\%$ and XGBoost at $81.6\%$), proving that incorporating semantic privileged information effectively bridges the gap between data-driven models and semantic reasoning agents.

In the lane-changing scenario, LUSPI demonstrates superior generalization capabilities, securing the best performance across all style categories with an average $F_1$-score of $91.0\% \pm 0.8\%$. This represents a performance gain of approximately $7.8\%$ over the SVM baseline ($84.4\% \pm 1.9\%$) and a distinct advantage over CatBoost ($87.5\% \pm 1.4\%$), the top-performing baseline in this domain. Notably, LUSPI achieves an $F_1$-score of $91.5\% \pm 1.0\%$ for ``Aggressive'' lane-changing maneuvers. Given that aggressive lane changes are critical precursors to traffic accidents, this high-precision recognition holds significant practical value for enhancing the safety warning capabilities of Advanced Driver Assistance Systems (ADAS). Furthermore, regarding the standard deviation, LUSPI exhibits minimal fluctuation across five-fold cross-validation ($1.4\%$ for car-following and $0.8\%$ for lane-changing), further corroborating the framework's stability.

To further substantiate the superiority of the proposed method, we conducted a statistical significance analysis. As indicated in the footnotes of Table \ref{tab:Performance-Combined}, paired $t$-test results confirm that the improvements of LUSPI in terms of overall average performance are statistically significant ($p < 0.05$) compared to all comparative methods. These findings suggest that LUSPI transcends mere data fitting; by integrating semantic reasoning knowledge generated by DriBehavGPT, it effectively overcomes the performance bottlenecks inherent in traditional methods that rely solely on kinematic features.

\begin{figure}[t]
    \centering
    \includegraphics[width=0.9\linewidth]{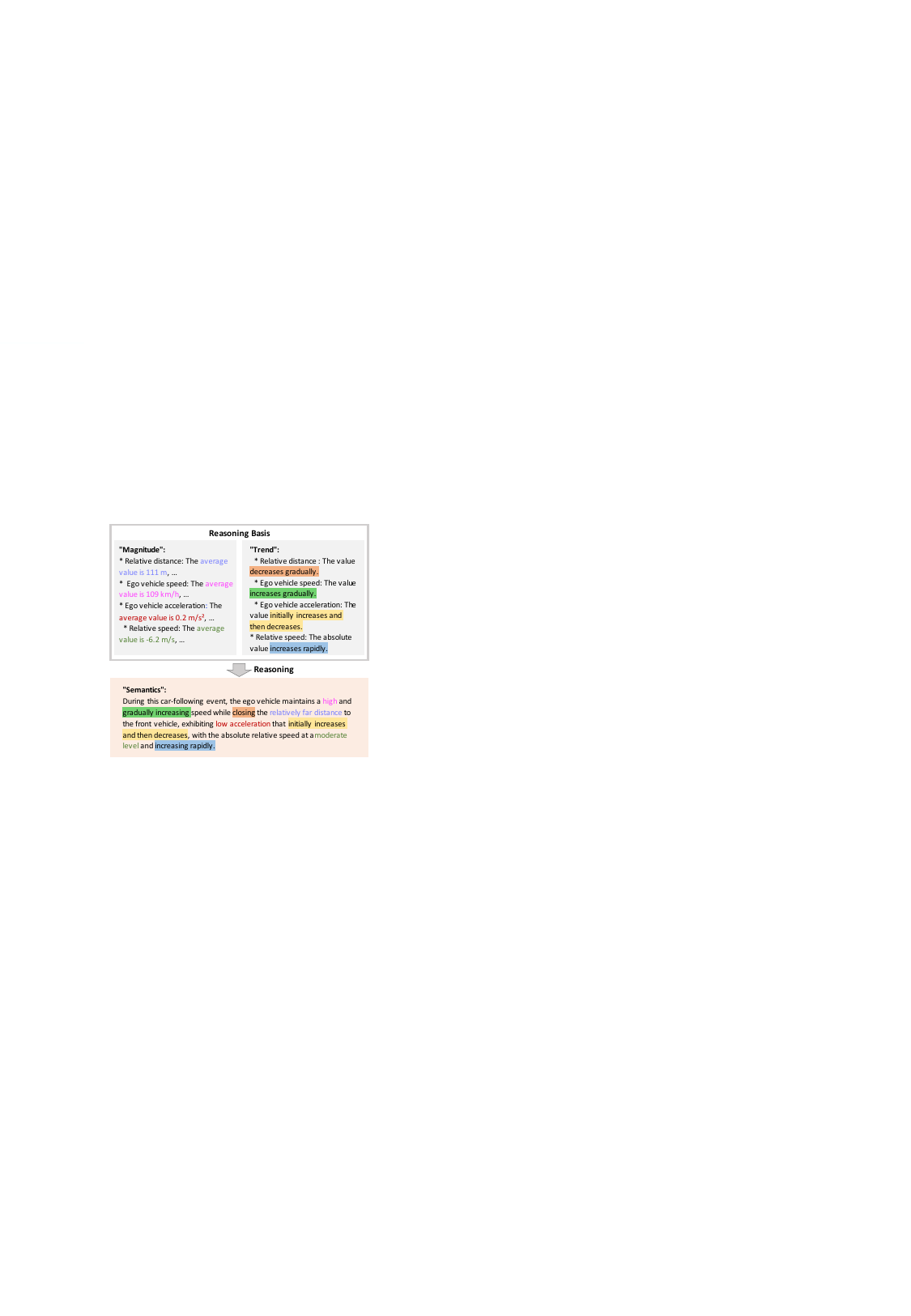}
    \caption{Illustration of chain-of-thought reasoning by  DriBehavGPT in a car-following scenario.}
    \label{fig: FlowLabeling1}
\end{figure}

\subsection{Interpretability of  DriBehavGPT}
The key advantage of DriBehavGPT lies in its structured analytical framework, which generates semantic, interpretable, and traceable outputs through step-by-step reasoning.  By decomposing complex driving decisions into human-understandable logical chains, the model provides transparent and explainable behavior analysis, effectively bridging the gap between opaque models and user cognition.

\subsubsection{Car-following Scenario}

DriBehavGPT generates interpretable explanations in car-following scenarios by combining magnitude and trend analysis. As illustrated in Fig. \ref{fig: FlowLabeling1}, the model’s semantic outputs are both descriptive and quantifiably grounded. For magnitude analysis, it captures static behavioral traits --- e.g., ``\textsf{\footnotesize \colorBlue{relatively far distance} to the front vehicle}'' is validated by ``\textsf{\footnotesize \colorBlue{average relative distance of $111$ m}}'', and ``\textsf{\footnotesize \colorPink{high-speed} travel}'' is supported by ``\textsf{\footnotesize \colorPink{average speed of $109$ km/h}}''. For trend analysis, it underpins dynamic behavioral changes --- e.g., ``\textsf{\footnotesize relative distance \hlfigorange{gradually decreasing}}'' indicates \textsf{\footnotesize \hlfigorange{closing}} proximity to the leading vehicle and ``\textsf{\footnotesize speed \hlfiggreen{gradually increasing}}'' suggests progressive acceleration. These analyses jointly enrich feature interpretation --- e.g., a negative relative speed with an increasing absolute value implies the ego vehicle is accelerating faster than the front vehicle, reinforcing the semantic explanation. This dual-faceted reasoning ensures outputs are not only human-readable but also rigorously justified by measurable driving dynamics.

\subsubsection{Lane-changing Scenario}

\begin{figure}[t]
    \centering
    \includegraphics[width=0.9\linewidth]{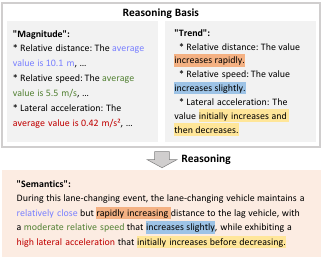}
    \caption{Illustration of chain-of-thought reasoning by  DriBehavGPT in a lane-changing scenario.}
    \label{fig:FlowLabeling-LC}
\end{figure}

The interpretability of LUSPI is further demonstrated in lane-changing scenarios (Fig.~\ref{fig:FlowLabeling-LC}), where magnitude and trend analyses provide complementary insights into driving behavior semantics. Magnitude analysis quantitatively grounds behavioral interpretations—for instance, the observation that the lane-changing vehicle ``\textsf{\footnotesize maintains a \colorBlue{relatively close} distance from the lag vehicle}'' is empirically validated by ``\textsf{\footnotesize \colorBlue{average relative distance of $10.1$ m}}'', while ``\textsf{\footnotesize \colorGreen{relative speed of $5.5$ m/s}}'' confirms it is overtaking the lag vehicle. Trend analysis reveals dynamic behavioral evolution, identifying patterns such as ``\textsf{\footnotesize relative distance \hlfigorange{increases rapidly}}'' or ``\textsf{\footnotesize relative speed \hlfigblue{increases slightly}}''. The synergy of these analyses enables nuanced semantic inference. For example, a moderate initial relative speed coupled with a gradual upward trend suggests the lane-changing vehicle not only travels faster but is also progressively accelerating within a medium-risk range.

This Chain-of-Thought (CoT) reasoning enhances trust in model outputs by transparently linking raw data (e.g., sensor measurements) to high-level behavioral conclusions. Unlike traditional opaque classifiers, LUSPI leverages the LLM's knowledge base and contextual reasoning to articulate decision pathways explicitly. This integration achieves superior interpretability without compromising performance—a critical advancement for safety-sensitive applications in autonomous driving.

\subsection{Impact of Dimensionality Reduction}

To evaluate the necessity of the dimensionality reduction module within our framework, we conducted an ablation study focusing on the processing of semantic privileged information. The text embeddings generated via SentenceTransformers inherently possess a high dimensionality of $D=768$. We investigated the effect of this high dimensionality on the learning process by comparing two experimental settings: (1) Raw embeddings, where the original 768-dimensional vectors are directly input into the SVM+ privileged kernel; and (2) Reduced embeddings (Ours), where features are projected into a compact latent space ($D=5$) via UMAP.

The results in Table \ref{tab:ablation_reduction} show a clear performance decline without dimensionality reduction: average $F_1$-scores drop by 3.4\% in car-following and 3.6\% in lane-changing. This performance degradation is likely attributed to the ``curse of dimensionality.'' The SVM+ algorithm relies on a kernel function (Eq. (3)) to measure the similarity between privileged data. Directly computing the kernel matrix in a 768-dimensional space renders distance measures less discriminative, causing the model to overfit the noise in the privileged information rather than capturing the underlying semantic structure. By reducing the dimensions to 5, we distill the core semantic information into a compact manifold, which provides a more robust regularization constraint for the optimization objective.

\begin{table}[htp]
\centering
\renewcommand{\arraystretch}{1.3}
\setlength{\tabcolsep}{4pt} 
\caption{Impact of Dimensionality Reduction on Average $F_1$-score (\%)}
\label{tab:ablation_reduction}
\begin{tabular}{lccc@{}} 
\hline\hline
Input & Dimension ($D$) & Car-following & Lane-changing \\
\hline
Raw Embeddings          & 768 & $86.8 \pm 1.8$ & $87.4 \pm 2.1$ \\
Reduced (Ours) & 5   & $90.2 \pm 1.4$ & $91.0 \pm 0.8$ \\
\hline\hline
\end{tabular}
\end{table}


\section{Conclusions}
\label{sec: Conclusions}
This paper presented a novel framework for driving style recognition that leverages sequential driving data augmented with semantic privileged information. The proposed method incorporates DriBehavGPT, an LLM-driven module, to generate human-interpretable descriptions of driving behavior. These semantic features serve as privileged information during training via SVM+, while only standard in-vehicle signals are required for inference, effectively addressing the limitation of conventional approaches that rely on identical input modalities across both phases.

Experimental validation in car-following and lane-changing scenarios demonstrates significant performance gains. In car-following, the framework achieves an average $F_1$-score of $90.2\%$, consistently outperforming a comprehensive set of baselines. For lane-changing, it attains a $91.0\%$ average $F_1$-score, demonstrating superior recognition capability compared to comparative methods. These results underscore the efficacy of integrating semantic reasoning into driving behavior analysis, advancing the development of reliable and interpretable intelligent driving systems. Future work will investigate multimodal privileged information (e.g., visual/auditory cues) and extend the framework to urban mixed-traffic environments.

\appendices

\section{}
\label{sec: Appendix HC}

The Lagrangian function we construct is as follows:
\begin{equation}
\begin{aligned}
    L&(\mathbf{w},\mathbf{b},\boldsymbol{\alpha},\boldsymbol{\beta}) \\
    &= \frac{1}{2} \left(\|\mathbf{w}\|^2+\gamma\|{\mathbf{w}^*}\|^2\right) + C \sum_{i=1}^n \left( {\mathbf{w}^*}^{\top} \psi(\mathbf{x}^*_i)+b^* \right)\\
    &- \sum_{i=1}^{n}\alpha_i \left[ y_i\left(\mathbf{w}^{\top} \phi\left(\mathbf{x}_i\right)+b\right) -1    
    + \left({\mathbf{w}^*}^{\top} \psi(\mathbf{x}^*_i)+b^*\right)\right]\\ 
    &-\sum_{i=1}^{n}\beta_i\left[ {\mathbf{w}^*}^{\top} \psi(\mathbf{x}^*_i)+b^* \right]
\end{aligned}
\end{equation}
Following the principle of Lagrangian duality, we solve the optimization problem by first minimizing the Lagrangian function with respect to $(\mathbf{w}, b, \mathbf{w}^* , b^*)$, and then maximizing the resulting dual function with respect to the Lagrange multipliers $\boldsymbol{\alpha} = [\alpha_1,\dots,\alpha_n]^\top \in \mathbb{R}^n$ and $\boldsymbol{\beta} = [\beta_1,\dots,\beta_n]^\top \in \mathbb{R}^n$. 
\section{}
\label{sec: Appendix HC}
SMO algorithm iteratively maximizes the dual cost function by selecting the best maximally sparse feasible direction in each iteration and updating the corresponding $\alpha_{i}$ and $\beta_{i}$ such that the dual constraints are also satisfied. The analytical update procedure for Lagrange multipliers proceeds as follows:
\begin{equation}\alpha_i^\mathrm{new}=\alpha_i^\mathrm{old}+\eta\frac{\partial\mathcal{L}_d}{\partial\alpha_i}\end{equation}
\begin{equation}\beta_j^\mathrm{new}=\beta_j^\mathrm{old}+\eta\frac{\partial\mathcal{L}_d}{\partial\beta_j}\end{equation}
where $\eta$ is learning rate and $\mathcal{L}_d$ is dual objective function. The algorithm terminates its iterations when the Karush-Kuhn-Tucker (KKT) conditions are satisfied, which occurs when the gradient norm $\|\nabla\mathcal{L}_d\|$ of the dual objective function falls below a predefined threshold $\epsilon$, indicating functional convergence.

\section{}

\begin{table}[H] 
    \centering
    \setlength{\tabcolsep}{4pt} 
    \renewcommand{\arraystretch}{1.25} 
        \setcounter{table}{0}
    \renewcommand{\thetable}{C\arabic{table}}
    
    \caption{Statistical Analysis of Car-following Features (Mean $\pm$ SD)}\label{tab:Car_Following_Stats}
    
\begin{tabular}{@{} l l l l @{}} 
        \hline\hline
        Feature & Conservative & Moderate & Aggressive \\ 
        \hline
       
        Ego Speed ($\mathrm{m/s}$) & 
        $22.5 \pm 2.2$ & $27.8 \pm 3.5$ & $31.5 \pm 5.8$ \\

        Longitudinal Accel. ($\mathrm{m/s^2}$) & 
        $0.35 \pm 0.12$ & $0.62 \pm 0.28$ & $1.15 \pm 0.65$ \\

        Time-to-Collision ($\mathrm{s}$) & 
        $8.5 \pm 3.2$ & $5.2 \pm 2.1$ & $1.9 \pm 0.9$ \\
        \hline\hline
    \end{tabular}
\end{table}

\begin{table}[H] 
    \centering
    \setlength{\tabcolsep}{6pt} 
    \renewcommand{\arraystretch}{1.25} 
    
    \setcounter{table}{1}
    \renewcommand{\thetable}{C\arabic{table}}
    
    \caption{Statistical Analysis of Lane-changing Features (Mean $\pm$ SD)}\label{tab:Appendix_LC_Stats}
    
    \begin{tabular}{@{} l l l l @{}} 
        \hline\hline
        Feature & Conservative & Moderate & Aggressive \\ 
        \hline
        Relative Distance ($\mathrm{m}$) & 
        $58.7 \pm 7.0$ & $39.9 \pm 6.7$ & $24.9 \pm 9.9$ \\
        
        Relative Speed ($\mathrm{m/s}$) & 
        $8.8 \pm 2.8$ & $6.2 \pm 2.8$ & $4.6 \pm 4.6$ \\
        
        Lateral Accel. ($\mathrm{m/s^2}$) & 
        $0.35 \pm 0.1$ & $0.6 \pm 0.15$ & $0.95 \pm 0.3$ \\
        \hline\hline
    \end{tabular}
\end{table}

\section{}

\begin{table}[H] 
    \centering
    \setlength{\tabcolsep}{1pt} 
    \renewcommand{\arraystretch}{1.25} 
    
    \setcounter{table}{0}
    \renewcommand{\thetable}{D\arabic{table}}
    
    \caption{Optimal Hyperparameter Settings for the Baseline Algorithms in Car-following Scenario}\label{tab:Appendix_Parameters_Final}
    
    \begin{tabular}{@{} l >{\raggedright\arraybackslash}p{0.8\linewidth} @{}} 
        \hline\hline
        Name & Selected Value \\ 
        \hline
        KNN & 
        n\_neighbors = 9; weights = `distance'; P = 1 \\
        
        SVM & 
        C = 8; kernel = `rbf'; gamma = `scale' \\
        
        Random Forest & 
        n\_estimators = 300; max\_depth = 12; min\_samples\_split = 6 \\
        
        XGBoost & 
        n\_estimators = 300; max\_depth = 4; learning\_rate = 0.1 \\
        
        LightGBM & 
        n\_estimators = 300; num\_leaves = 15;
        learning\_rate = 0.05
        \\
        
        CatBoost & 
         iterations = 500; max\_depth = 5;
         learning\_rate = 0.1
         \\
        
        MLP & 
        hidden\_layer\_sizes = (64); solver = `adam'; learning\_rate = `adaptive' \\
        \hline\hline
    \end{tabular}
\end{table}

\begin{table}[H] 
    \centering
    \setlength{\tabcolsep}{1pt} 
    \renewcommand{\arraystretch}{1.25} 
    
    \setcounter{table}{1} 
    \renewcommand{\thetable}{D\arabic{table}}
    
    \caption{Optimal Hyperparameter Settings for the Baseline Algorithms in Lane-changing Scenario}\label{tab:Appendix_Parameters_CF}
    
    \begin{tabular}{@{} l >{\raggedright\arraybackslash}p{0.8\linewidth} @{}} 
        \hline\hline
        Name & Selected Value \\ 
        \hline
        KNN & 
        n\_neighbors = 11; weights = `distance'; P = 1 \\
        
        SVM & 
        C = 10; kernel = `rbf'; gamma = `scale' \\
        
        Random Forest & 
        n\_estimators = 500; max\_depth = 10; min\_samples\_split = 4 \\
        
        XGBoost & 
        n\_estimators = 400; max\_depth = 5; learning\_rate = 0.05 \\
        
        LightGBM & 
        n\_estimators = 500; num\_leaves = 12;
        learning\_rate = 0.05
        \\
        
        CatBoost & 
         iterations = 500; max\_depth = 6;
         learning\_rate = 0.1
         \\
        
        MLP & 
        hidden\_layer\_sizes = (64); solver = `adam'; learning\_rate = `adaptive' \\
        \hline\hline
    \end{tabular}
\end{table}




\bibliographystyle{IEEEtran}
\bibliography{refs}

\ifCLASSOPTIONcaptionsoff
  \newpage
\fi

\begin{IEEEbiography}[{\includegraphics[width=1in,height=1.25in,clip,keepaspectratio]{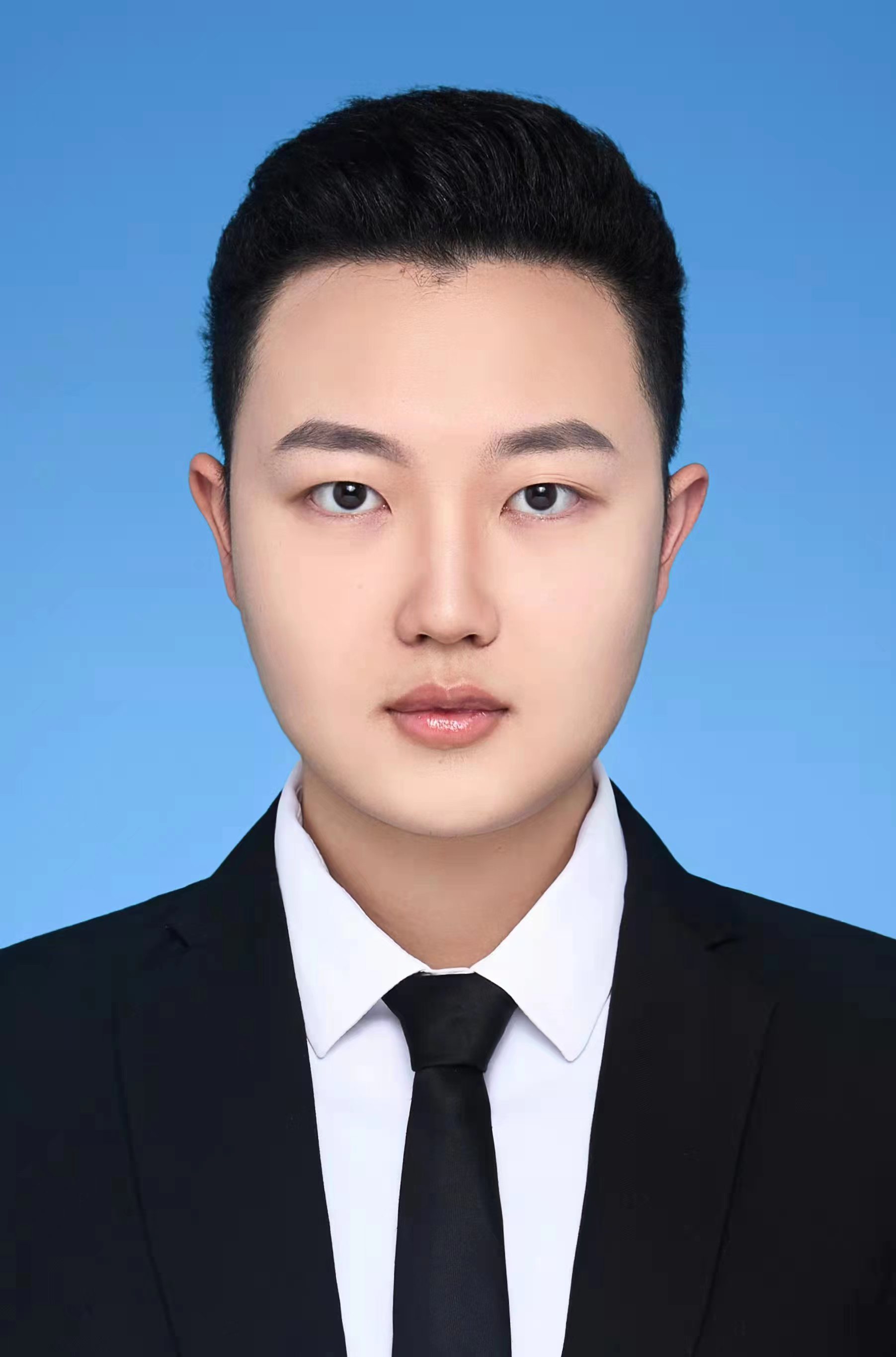}}]{Zhaokun Chen} received the B.S. degree in Intelligent Automotive Engineering from the Harbin Institute of Technology, Weihai, China, in 2022. He is currently working toward the Ph.D. degree in mechanical engineering with the Beijing Institute of Technology, Beijing, China. His research interests include driving behavior analysis, human-robot interaction, human intent recognition, and shared control.
\end{IEEEbiography}

\begin{IEEEbiography}[{\includegraphics[width=1in,height=1.25in,clip,keepaspectratio]{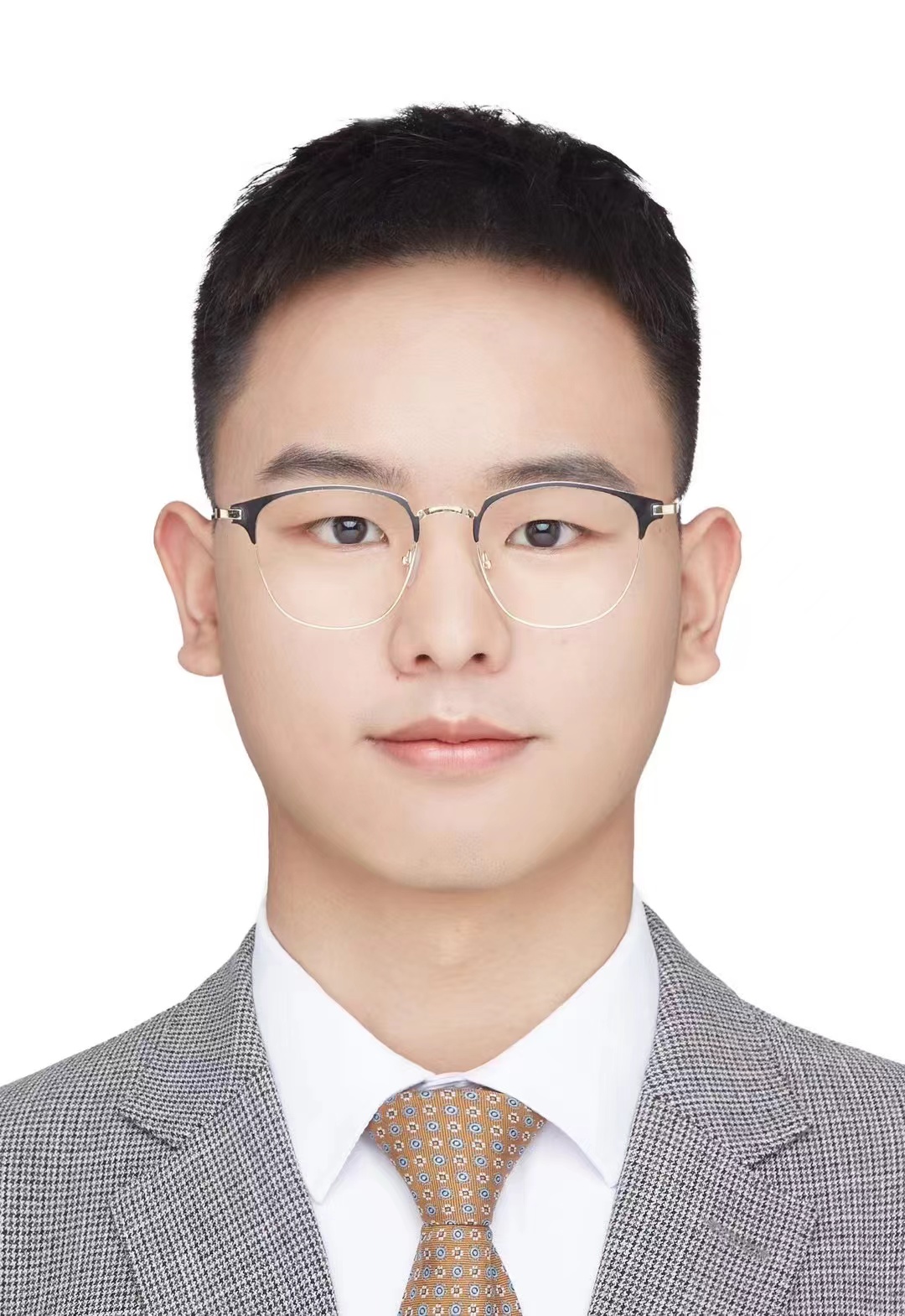}}]{Chaopeng Zhang} received B.S. degree in Mechanical Engineering from Beijing Institute of Technology, Beijing, China, in 2019, where he is currently working toward the Ph.D. degree in mechanical engineering. 
He is now studying at the Department of Mechanical Engineering, University of Tokyo. His research interests include human factors in intelligent vehicles, human driver models, driving style recognition, and driving intention recognition.
\end{IEEEbiography}

\begin{IEEEbiography}[{\includegraphics[width=1in,height=1.25in,clip,keepaspectratio]{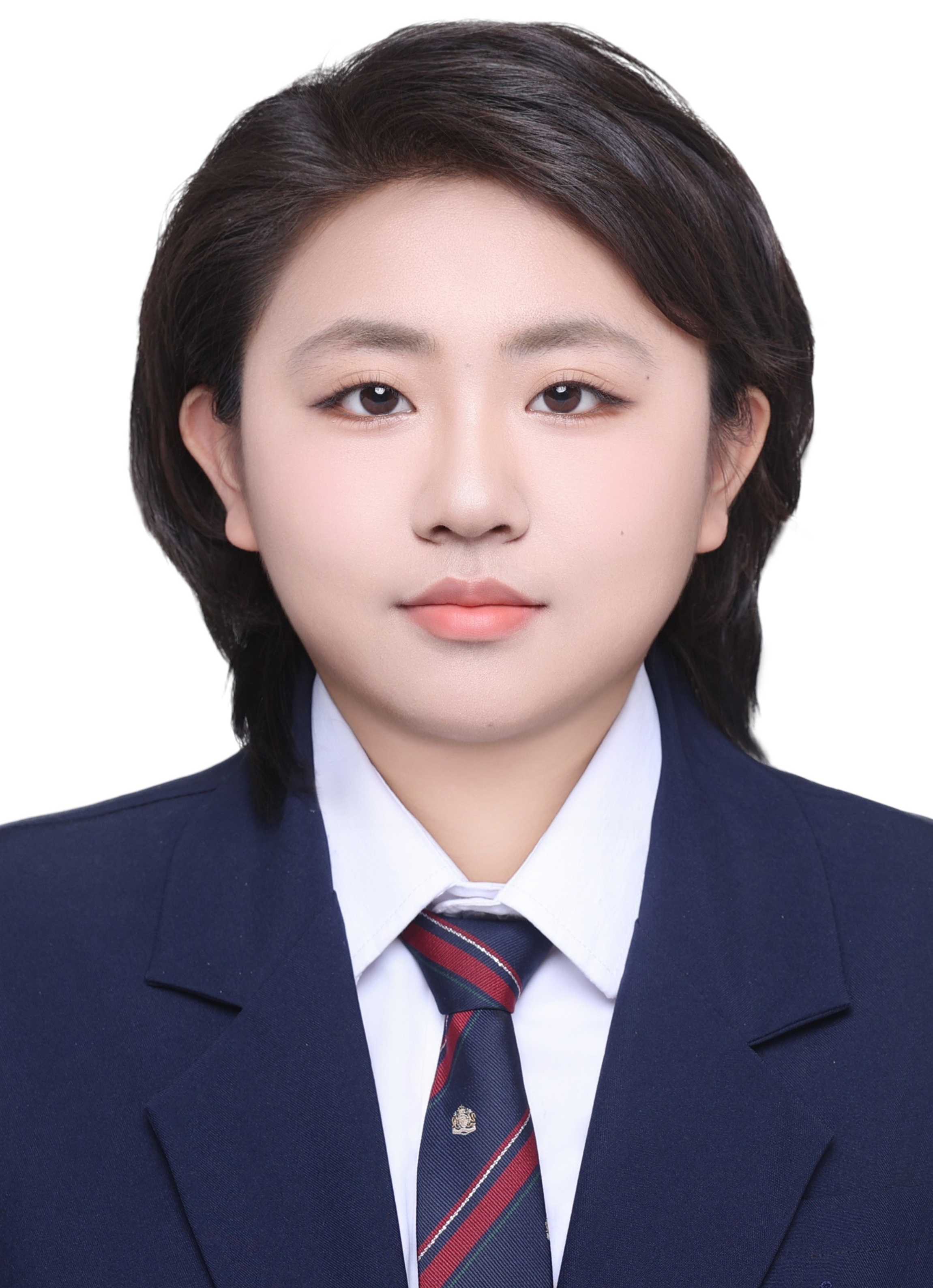}}]{Xiaohan Li} received the B.S. degree in Automotive Engineering from Jilin University, Changchun, China, in 2025. She is currently working toward the M.S. degree in mechanical engineering with the Beijing Institute of Technology, Beijing, China. Her research interests include driving style recognition, human factors in intelligent vehicles, and human intention recognition.
\end{IEEEbiography}

\begin{IEEEbiography}[{\includegraphics[width=1in,height=1.25in,clip,keepaspectratio]{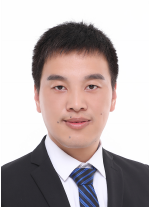}}]{Wenshuo Wang} (SM'15-M'18) received his Ph.D. degree in mechanical engineering from the Beijing Institute of Technology (BIT) in 2018.  Presently, he is a Full Professor at the School of Mechanical Engineering, BIT, Beijing, China. Prior to his role at BIT, he completed Postdoctoral fellowships at McGill University, Carnegie Mellon University (CMU), and UC Berkeley between 2018 and 2023. Furthermore, from 2015 to 2018, he served as a Research Assistant at UC Berkeley and the University of Michigan, Ann Arbor. His research interests focus on Bayesian nonparametric learning, human driver model, human–vehicle interaction, ADAS, and autonomous vehicles.
\end{IEEEbiography}

\begin{IEEEbiography}[{\includegraphics[width=1in,height=1.25in,clip,keepaspectratio]{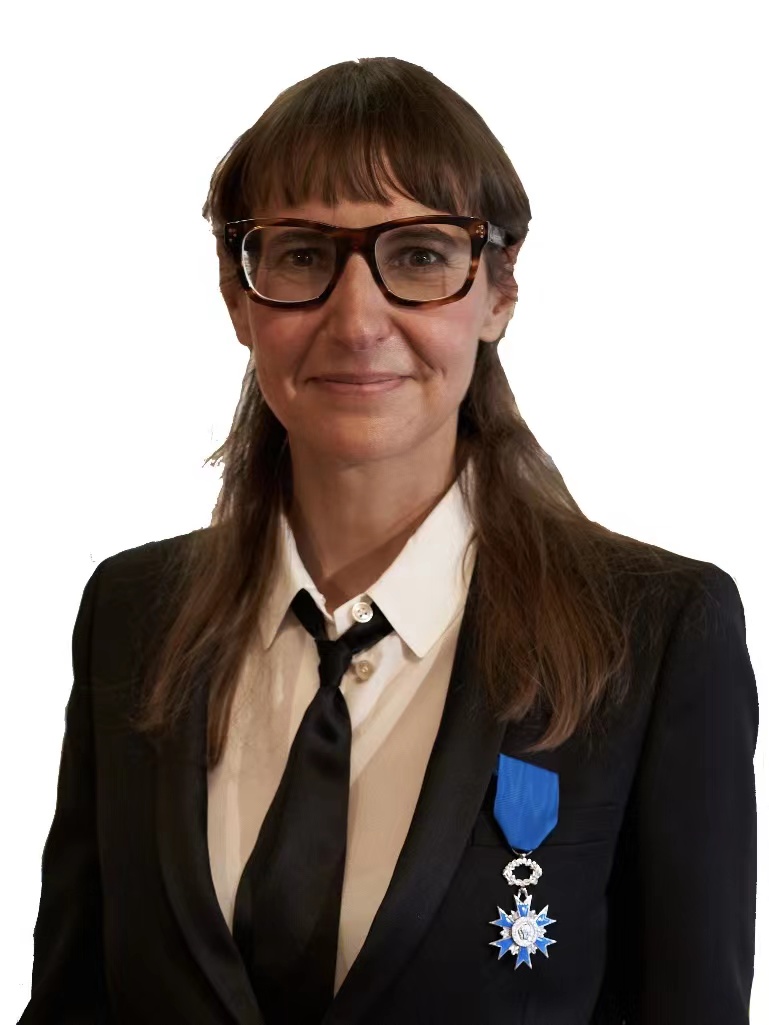}}]{Gentiane Venture} is a French Roboticist working in academia in Tokyo. She is a professor at the University of Tokyo and a cross-appointed fellow with AIST. She obtained her MSc and PhD from Ecole Centrale/University of Nantes in 2000 and 2003 respectively. She worked at CEA, France in 2004 and for 5 years at the University of Tokyo, Japan. In 2009 she started with Tokyo University of Agriculture and Technology where she established an international research group working on human science and robotics, before moving to her present affiliation in 2022. With her group, she conducts theoretical and applied research on motion dynamics, robot control, and non-verbal communication to study the meaning of living with robots. Her work is highly interdisciplinary, collaborating with therapists, psychologists, neuroscientists, sociologists, philosophers, ergonomists, artists, and designers.
\end{IEEEbiography}

\begin{IEEEbiography}[{\includegraphics[width=1in,height=1.25in,clip,keepaspectratio]{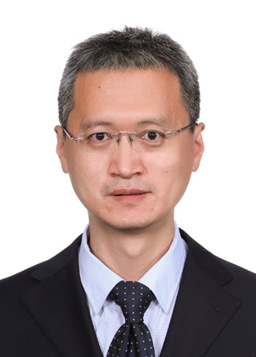}}]{Junqiang Xi} received the B.S. degree in automotive engineering from the Harbin Institute of Technology, Harbin, China, in 1995, and the Ph.D. degree in vehicle engineering from the Beijing Institute of Technology (BIT), Beijing, China, in 2001. In 2001, he joined the State Key Laboratory of Vehicle Transmission, BIT. During 2012–2013, he made research as an Advanced Research Scholar in Vehicle Dynamic and Control Laboratory, Ohio State University, Columbus, OH, USA. He is currently a Professor and Director of Automotive Research Center in BIT. His research interests include vehicle dynamic and control, powertrain control, mechanics, intelligent transportation system, and intelligent vehicles.
\end{IEEEbiography}



\end{document}